%% file: main.tex
\documentclass[remotesensing,technicalnote,accept,pdftex,moreauthors]{Definitions/mdpi} 

\input{Definitions/package}

\include{Definitions/unicode}

\firstpage{1} 
\pubvolume{15}
\issuenum{12}
\articlenumber{2970}
\pubyear{2023}
\copyrightyear{2023}
\externaleditor{Academic Editors: Fabio Remondino and Rongjun Qin}
\datereceived{19 April 2023}
\daterevised{3 June 2023}
\dateaccepted{5 June 2023}
\datepublished{7 June 2023}

\hreflink{https://doi.org/\linebreak10.3390/rs15122970} 

\pdfoutput=1

\usepackage{epsfig}
\usepackage{graphicx}
\usepackage{amsmath}
\usepackage{amssymb}
\usepackage{widetable}

\usepackage[labelformat=simple]{subfig}

\usepackage{placeins}
\usepackage{svg}

\Title{DDL-MVS: Depth Discontinuity Learning for Multi-View Stereo Networks}

\TitleCitation{DDL-MVS: Depth Discontinuity Learning for Multi-View Stereo Networks}

\Author{Nail 
 Ibrahimli \textsuperscript{1}\textsuperscript{,}*
\orcidA{}, Hugo Ledoux \textsuperscript{1}\orcidB{}, Julian F. P. Kooij \textsuperscript{2}\orcidC{} and Liangliang Nan  \textsuperscript{1}\orcidD{}}

\AuthorNames{Nail Ibrahimli, Hugo Ledoux, Julian F. P. Kooij and Liangliang Nan}

\AuthorCitation{Ibrahimli, 
 N.; Ledoux, H.; Kooij, J.F.P.; Nan, L.}

\address{$^{1}$ \quad 3D Geoinformation, Faculty of Architecture and the Built Environment, Delft University of Technology,\linebreak 2628 BL Delft, The Netherlands; liangliang.nan@tudelft.nl (L.N.)\\ 
$^{2}$ \quad Intelligent Vehicles Group, Delft University of Technology, 2628 BL Delft, The Netherlands}

\corres{Correspondence: n.ibrahimli@tudelft.nl}

\abstract{We propose an enhancement module called depth discontinuity learning (DDL) for learning-based  multi-view  stereo (MVS) methods. 
Traditional methods are known for their accuracy but struggle with completeness.
While recent learning-based methods have improved completeness at the cost of accuracy, our DDL approach aims to improve accuracy while retaining completeness in the reconstruction process.
To achieve this, we introduce the joint estimation of depth and boundary maps, where the boundary maps are explicitly utilized for further refinement of the depth maps.
We validate our idea by integrating it into an existing learning-based MVS pipeline where the reconstruction depends on high-quality depth map estimation. 
Extensive experiments on various datasets, namely DTU,  ETH3D, ``Tanks and Temples'', and BlendedMVS, show that our method improves reconstruction quality compared to our baseline, Patchmatchnet. 
Our ablation study demonstrates that incorporating the proposed DDL  significantly reduces the depth map error, for instance, by more than 30\% on the DTU dataset, and leads to improved depth map quality in both smooth and boundary regions. Additionally, our qualitative analysis has shown that the reconstructed point cloud exhibits enhanced quality without any significant compromise on completeness.
Finally, the experiments reveal that our proposed model and strategies exhibit strong generalization capabilities across the various datasets.}

\begin{document}

\input{source/introduction}
\input{source/related_work}

\input{source/method}

\input{source/experiments}

\input{source/conclusion}

\vspace{6pt}


    


\begin{adjustwidth}{-\extralength}{0cm}

\reftitle{References}


\bibliography{egbib}

\end{adjustwidth}
\end{document}

%% file: Definitions/package.tex
\usepackage[normalem]{ulem} 
\usepackage{amsfonts} 
\usepackage{wasysym} 
\usepackage{mathcomp} 
\usepackage{CJKutf8} 
\usepackage{pifont} 
\usepackage{bm} 
\graphicspath{{./Definitions/}} 

\makeatletter
\def\T@n@@nc@d@ngM@cr@M@d{}
\def\LY@n@@nc@d@ngM@cr@M@d{}
\makeatother

\let\orignewcommand\newcommand  
\let\newcommand\providecommand  
\usepackage{verse}
\let\newcommand\orignewcommand  

\newsavebox\foobox

\renewcommand{\dagger}{\mathchar"2279}
\renewcommand{\ddagger}{\mathchar"227A}

\newcommand{\mmathit}[1]{
  \ifthenelse{\equal{#1}{\ln}}{\mathit{ln}}{
    \ifthenelse{\equal{#1}{\max}}{\mathit{max}}{\mathit{#1}}
  }
}
\makeatother


%% file: Definitions/unicode.tex
\DeclareUnicodeCharacter{1E45}{\.{n}}
\DeclareUnicodeCharacter{1E41}{\.{m}}
\DeclareUnicodeCharacter{2003}{\quad}
\DeclareUnicodeCharacter{2009}{\thinspace}
\DeclareUnicodeCharacter{2002}{\enspace{}}
\DeclareUnicodeCharacter{2005}{\thinspace}
\DeclareUnicodeCharacter{0263}{\textipa{G}}
\DeclareUnicodeCharacter{A0}{~}
\DeclareUnicodeCharacter{2460}{\textcircled{\scriptsize{1}}}
\DeclareUnicodeCharacter{2461}{\textcircled{\scriptsize{2}}}
\DeclareUnicodeCharacter{2462}{\textcircled{\scriptsize{3}}}
\DeclareUnicodeCharacter{2463}{\textcircled{\scriptsize{4}}}
\DeclareUnicodeCharacter{2464}{\textcircled{\scriptsize{5}}}
\DeclareUnicodeCharacter{2465}{\textcircled{\scriptsize{6}}}
\DeclareUnicodeCharacter{2466}{\textcircled{\scriptsize{7}}}
\DeclareUnicodeCharacter{2467}{\textcircled{\scriptsize{8}}}
\DeclareUnicodeCharacter{2468}{\textcircled{\scriptsize{9}}}
\DeclareUnicodeCharacter{2070}{\textsuperscript{0}}
\DeclareUnicodeCharacter{2074}{\textsuperscript{4}}
\DeclareUnicodeCharacter{2075}{\textsuperscript{5}}
\DeclareUnicodeCharacter{2076}{\textsuperscript{6}}
\DeclareUnicodeCharacter{2077}{\textsuperscript{7}}
\DeclareUnicodeCharacter{2078}{\textsuperscript{8}}
\DeclareUnicodeCharacter{2079}{\textsuperscript{9}}
\DeclareUnicodeCharacter{02C2}{<}
\DeclareUnicodeCharacter{2033}{''}
\DeclareUnicodeCharacter{0229}{\c{e}}
\DeclareUnicodeCharacter{016F}{\r{u}}
\DeclareUnicodeCharacter{3AC}{\relax\ifmmode\acute{\alpha}\else $\acute{\alpha}$\fi}
\DeclareUnicodeCharacter{3AD}{\relax\ifmmode\acute{\varepsilon}\else $\acute{\varepsilon}$\fi}
\DeclareUnicodeCharacter{3AE}{\relax\ifmmode\acute{\eta}\else $\acute{\eta}$\fi}
\DeclareUnicodeCharacter{3AF}{\relax\ifmmode\acute{\iota}\else $\acute{\iota}$\fi}
\DeclareUnicodeCharacter{3CC}{\relax\ifmmode\acute{o}\else $\acute{o}$\fi}
\DeclareUnicodeCharacter{3CD}{\relax\ifmmode\acute{\upsilon}\else $\acute{\upsilon}$\fi}
\DeclareUnicodeCharacter{3CE}{\relax\ifmmode\acute{\omega}\else $\acute{\omega}$\fi}
\DeclareUnicodeCharacter{391}{A}
\DeclareUnicodeCharacter{392}{B}
\DeclareUnicodeCharacter{395}{E}
\DeclareUnicodeCharacter{396}{Z}
\DeclareUnicodeCharacter{397}{H}
\DeclareUnicodeCharacter{399}{I}
\DeclareUnicodeCharacter{39A}{K}
\DeclareUnicodeCharacter{39C}{M}
\DeclareUnicodeCharacter{39D}{N}
\DeclareUnicodeCharacter{39F}{O}
\DeclareUnicodeCharacter{3A1}{P}
\DeclareUnicodeCharacter{3A4}{T}
\DeclareUnicodeCharacter{3A7}{X}

\DeclareUnicodeCharacter{27E6}{\relax\ifmmode \llbracket \else $\llbracket$\fi}
\DeclareUnicodeCharacter{27E7}{\relax\ifmmode \rrbracket \else $\rrbracket$\fi}

\DeclareUnicodeCharacter{1D434}{\relax\ifmmode A \else $A$\fi}
\DeclareUnicodeCharacter{1D435}{\relax\ifmmode B \else $B$\fi}
\DeclareUnicodeCharacter{1D436}{\relax\ifmmode C \else $C$\fi}
\DeclareUnicodeCharacter{1D437}{\relax\ifmmode D \else $D$\fi}
\DeclareUnicodeCharacter{1D438}{\relax\ifmmode E \else $E$\fi}
\DeclareUnicodeCharacter{1D439}{\relax\ifmmode F \else $F$\fi}
\DeclareUnicodeCharacter{1D43A}{\relax\ifmmode G \else $G$\fi}
\DeclareUnicodeCharacter{1D43B}{\relax\ifmmode H \else $H$\fi}
\DeclareUnicodeCharacter{1D43C}{\relax\ifmmode I \else $I$\fi}
\DeclareUnicodeCharacter{1D43D}{\relax\ifmmode J \else $J$\fi}
\DeclareUnicodeCharacter{1D43E}{\relax\ifmmode K \else $K$\fi}
\DeclareUnicodeCharacter{1D43F}{\relax\ifmmode L \else $L$\fi}
\DeclareUnicodeCharacter{1D440}{\relax\ifmmode M \else $M$\fi}
\DeclareUnicodeCharacter{1D441}{\relax\ifmmode N \else $N$\fi}
\DeclareUnicodeCharacter{1D442}{\relax\ifmmode O \else $O$\fi}
\DeclareUnicodeCharacter{1D443}{\relax\ifmmode P \else $P$\fi}
\DeclareUnicodeCharacter{1D444}{\relax\ifmmode Q \else $Q$\fi}
\DeclareUnicodeCharacter{1D445}{\relax\ifmmode R \else $R$\fi}
\DeclareUnicodeCharacter{1D446}{\relax\ifmmode S \else $S$\fi}
\DeclareUnicodeCharacter{1D447}{\relax\ifmmode T \else $T$\fi}
\DeclareUnicodeCharacter{1D448}{\relax\ifmmode U \else $U$\fi}
\DeclareUnicodeCharacter{1D449}{\relax\ifmmode V \else $V$\fi}
\DeclareUnicodeCharacter{1D44A}{\relax\ifmmode W \else $W$\fi}
\DeclareUnicodeCharacter{1D44B}{\relax\ifmmode X \else $X$\fi}
\DeclareUnicodeCharacter{1D44C}{\relax\ifmmode Y \else $Y$\fi}
\DeclareUnicodeCharacter{1D44D}{\relax\ifmmode Z \else $Z$\fi}
\DeclareUnicodeCharacter{1D44E}{\relax\ifmmode a \else $a$\fi}
\DeclareUnicodeCharacter{1D44F}{\relax\ifmmode b \else $b$\fi}
\DeclareUnicodeCharacter{1D450}{\relax\ifmmode c \else $c$\fi}
\DeclareUnicodeCharacter{1D451}{\relax\ifmmode d \else $d$\fi}
\DeclareUnicodeCharacter{1D452}{\relax\ifmmode e \else $e$\fi}
\DeclareUnicodeCharacter{1D453}{\relax\ifmmode f \else $f$\fi}
\DeclareUnicodeCharacter{1D454}{\relax\ifmmode g \else $g$\fi}
\DeclareUnicodeCharacter{1D456}{\relax\ifmmode i \else $i$\fi}
\DeclareUnicodeCharacter{1D457}{\relax\ifmmode j \else $j$\fi}
\DeclareUnicodeCharacter{1D458}{\relax\ifmmode k \else $k$\fi}
\DeclareUnicodeCharacter{1D459}{\relax\ifmmode l \else $l$\fi}
\DeclareUnicodeCharacter{1D45A}{\relax\ifmmode m \else $m$\fi}
\DeclareUnicodeCharacter{1D45B}{\relax\ifmmode n \else $n$\fi}
\DeclareUnicodeCharacter{1D45C}{\relax\ifmmode o \else $o$\fi}
\DeclareUnicodeCharacter{1D45D}{\relax\ifmmode p \else $p$\fi}
\DeclareUnicodeCharacter{1D45E}{\relax\ifmmode q \else $q$\fi}
\DeclareUnicodeCharacter{1D45F}{\relax\ifmmode r \else $r$\fi}
\DeclareUnicodeCharacter{1D460}{\relax\ifmmode s \else $s$\fi}
\DeclareUnicodeCharacter{1D461}{\relax\ifmmode t \else $t$\fi}
\DeclareUnicodeCharacter{1D462}{\relax\ifmmode u \else $u$\fi}
\DeclareUnicodeCharacter{1D463}{\relax\ifmmode v \else $v$\fi}
\DeclareUnicodeCharacter{1D464}{\relax\ifmmode w \else $w$\fi}
\DeclareUnicodeCharacter{1D465}{\relax\ifmmode x \else $x$\fi}
\DeclareUnicodeCharacter{1D466}{\relax\ifmmode y \else $y$\fi}
\DeclareUnicodeCharacter{1D467}{\relax\ifmmode z \else $z$\fi}

\DeclareUnicodeCharacter{1E67}{\.{\v s}}
\DeclareUnicodeCharacter{1E11}{\relax\ifmmode \c{d} \else $\c{d}$\fi}
\DeclareUnicodeCharacter{1ECB}{\relax\ifmmode \d{i} \else $\d{i}$\fi}
\DeclareUnicodeCharacter{1D8D}{\relax\ifmmode \textlhookx \else $\textlhookx$\fi}
\DeclareUnicodeCharacter{104}{\relax\ifmmode \k{A} \else $\k{A}$\fi}
\DeclareUnicodeCharacter{211E}{\relax\ifmmode \textrecipe \else $\textrecipe$\fi}
\DeclareUnicodeCharacter{29D}{\relax\ifmmode \textctj \else $\textctj$\fi}

\DeclareUnicodeCharacter{1E2E}{\'{\"I}}
\DeclareUnicodeCharacter{23F}{\textrts}

\DeclareUnicodeCharacter{2C73}{\varw}

\DeclareUnicodeCharacter{2127}{\mho}

\DeclareUnicodeCharacter{28C}{\textturnv}
\DeclareUnicodeCharacter{252}{\textturnscripta}
\DeclareUnicodeCharacter{259}{\schwa}
\DeclareUnicodeCharacter{25B}{\m{e}}
\DeclareUnicodeCharacter{266}{\m{h}}
\DeclareUnicodeCharacter{127}{\B{h}}
\DeclareUnicodeCharacter{27E}{\textfishhookr}
\DeclareUnicodeCharacter{281}{\textinvscr}

%% file: source/introduction.tex
\section{Introduction}%
\label{sec:intro}

Multi-view stereo (MVS) techniques have been widely used to obtain dense 3D reconstruction from images. 
 MVS allows aerial images to be converted into accurate 3D models, which provide a more comprehensive representation of the large scene. This 3D information can be used for various applications, such as digital surface modelling~\cite{lemaire2008aspects}, landform analysis~\cite{peppa2019automated}, and urban planning~\cite{Nguatem_2017_ICCV}. It provides valuable insights into the shape, structure, and topography of the scene, enabling better understanding and interpretation of remote sensing data.

Traditional MVS techniques \cite{furukawa2010, galliani_2015_gipuma, tola2012efficient} extract dense correspondences from multiple calibrated views and generate a dense 3D representation (i.e., point cloud or dense triangle mesh) of the scene. 
These methods rely on image correspondences in the RGB space, which are sensitive to textureless and non-Lambertian surfaces, and lighting variations.
Recent developments in deep learning allow the use of learned feature maps instead of directly working on RGB images to build more robust MVS pipelines~\cite{yao_2018_mvsnet,yao_2019_rmvsnet,ji_2017_surfacenet,chen_2019_pointmvsnet,Yu_2020_fastmvsnet,cheng_2020_ucsnet,gu_2020_cascademvsnet,luo_2019_p_mvsnet,xu2020learning_inverse,yang_2020_cvpr,wang2020patchmatchnet}. 
By learning feature maps about the objects in the scene, learning-based MVS methods have demonstrated better completeness than traditional methods in reconstructing man-made objects with low texture and non-Lambertian surfaces. 
Recent learning-based MVS methods learn to reconstruct the depth map from input images by regularizing the 3D cost volume~\cite{yao_2018_mvsnet,gu_2020_cascademvsnet} or by Patchmatch-based iterative optimization~\cite{wang2020patchmatchnet,Duggal2019ICCV}.
Still, depth estimation remains challenging, and depth discontinuities at transitions between object boundaries are  usually erroneous~\cite{zhu2020edge,Tosi2021CVPR}. 
While this kind of error can be  alleviated by post-processing filters, it often reduces the completeness of the reconstruction.

\begin{figure}[t]
		\centering
		\includegraphics
		[width=0.7\columnwidth]{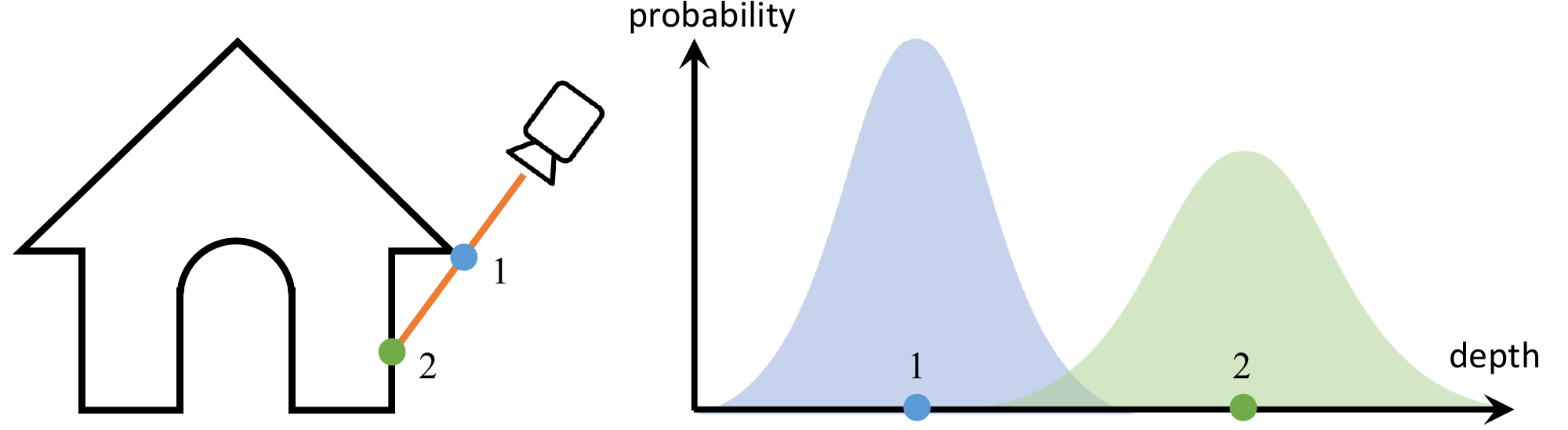}
		\caption{We propose to estimate depth as a bimodal univariate distribution. Using this depth representation, we improve multi-view depth reconstruction, especially across geometric boundaries.}
		\label{fig:bimodal_depth}
	\end{figure}

In MVS pipelines, it is common for a single depth value to be estimated per pixel, accompanied by a smooth surface assumption. This spatial regularization technique results in higher-quality depth maps, as shown in previous studies such as ~\cite{boykov2001fast,boykov_cvpr_1998}, which in turn improves the completeness of the reconstructed 3D model. However, a limitation of this approach is that it tends to oversmooth the true depth continuities at object boundaries, as pointed out in recent works such as \cite{garg2020wasserstein,Tosi2021CVPR}. Furthermore, as illustrated in Fig.\ref{fig:bimodal_depth}, pixels near depth discontinuities can pose ambiguity in determining which side of the depth boundary they belong to.


These findings motivate us to pursue two complementary objectives.
First, we aim to explicitly detect the geometric edges, instead of relying solely on photometric edges that capture color and texture changes~\cite{boykov2001fast,boykov_cvpr_1998}. 
Geometric edges more accurately indicate the true locations of object boundaries than photometric edges (see Fig.~\ref{fig:bimodal_depth}).  
We propose to estimate geometric boundary maps jointly with the depth maps, such that smooth depth surfaces can be enforced while considering the local geometry.
Second, as shown in  Fig.~\ref{fig:bimodal_depth}, we propose to estimate per-pixel depth as a univariate bimodal distribution rather than as a single depth value. 
This allows us to explicitly represent the depth ambiguity and avoids over-smoothing the depth discontinuities. 
We integrate both objectives into a multi-task learning architecture to improve the depth accuracy while avoiding the completeness trade-off of previous approaches.

To confirm the validity of our idea, we integrate it into the existing learning-based Multi-View Stereo (MVS) pipeline.
Extensive experiments that we ran on various benchmark datasets (see Sect.~\ref{sec:exp}) demonstrate that our method obtains better results.\footnote{The code is available at \url{https://github.com/mirmix/ddlmvs}}
Moreover, our method has high generalization capabilities, which have been validated by training our model on one dataset and testing it on other datasets.

In summary, the contributions of this work to multi-view stereo networks are: (1) a novel multi-task learning architecture for joint estimation of depth maps and object boundary maps for learning-based multi-view stereo pipelines; (2) a bimodal depth representation that represents depth as a distribution learned from multi-view images;  (3) a general loss formulation for depth discontinuity-based spatial regularization, which helps to learn discontinuities in depth and to regularize the depth maps.

The structure of this article is as follows: In Section 2, we will review the existing literature and contrast the most related works to our approach. Section 3 will delve into our methodology and outline the MVS pipeline we use to test our approach. Section 4 will present and discuss the experimental results obtained from our study. Finally, in Section 5, we will conclude the paper by summarizing our main findings.

%% file: source/related_work.tex
\section{Related work}
\label{sec:related_works}

As learning-based MVS networks are inspired by photogrammetry-based MVS algorithms and developed from two-view methods, we review photogrammetry-based MVS algorithms, learning-based two-view methods, and the recent development in learning-based MVS networks.

\subsection{Photogrammetry-based MVS} 
Multi-View stereo methods purely built upon photogrammetry and multi-view geometry theory are usually referred to as traditional multi-view stereo methods. Janai et al.~\cite{janai2020computer} showed that the taxonomy of the traditional multi-view stereo methods can be divided into four classes based on their representations of the scene and output. These scene representations are depth maps, point clouds, volumetric representations, and mesh or surfaces. 

Volumetric representations use either discrete occupancy function~\cite{kutulakos2000theory} or levelset alike signed distance functions~\cite{faugeras2002variational}, which limits them to small-scale reconstruction. 
The most common mesh-based approaches run variations of the marching cubes algorithm~\cite{lorensen1987marching} on top of a signed distance function based on a volumetric surface representation~\cite{curless1996volumetric}. 

The seminal point cloud-based method by Furukawa et.al.~\cite{furukawa2010} has shown that starting with an initial sparse set of point features it is possible to create an initial set of patches and densify them by iterative greedy expansion and photo-geometric filtering. These methods usually demand a uniformly sampled sparse set of points across the image domain to be able to create point clouds with better completeness.

Depth map-based approaches usually first try to estimate a 2.5D depth map for each view. By using multi-view fusion pipelines~\cite{zach2007globally,curless1996volumetric}, these depth maps are consolidated into a single geometric model. Although the plane sweeping algorithm~\cite{collins1996space} has high memory consumption, it was the most commonly used technique for depth map estimation. To use plane sweeping stereo for a large dynamic range of outdoor videos, Pollefeys et al.~\cite{pollefeys2008detailed}  took advantage of GPS and inertia measurements to place the reconstructed models in geo-registered coordinates. Using random initialization and propagation techniques, the PatchMatch-based MVS algorithms~\cite{galliani_2015_gipuma,schoenberger2016mvs} were able to estimate the depth map of each view with low memory consumption. In this work, we use a differentiable PatchMatch-based module to achieve a similar goal.

\subsection{Learning-based two-view methods}
Learning-based two-view methods have introduced the initial building blocks for two-view stereo matching and depth estimation, which were later adapted for multi-view settings. The most common building blocks for learning-based depth map estimation pipelines are feature extraction and depth estimation from the feature space. Shared weight-based feature extraction was introduced by~\cite{zbontar2016stereo}, and later improved by using cost volume regularization for depth map extraction~\cite{kendall2017end,chang2018pyramid,yang2019hierarchical}.
To reduce memory demand of the cost volume, Duggal et al.~\cite{Duggal2019ICCV} introduced differentiable PatchMatch Stereo (PMS) for two-view depth map estimation.
These approaches were later adapted for multi-view settings via differentiable homography~\cite{chang2018pyramid,cheng_2020_ucsnet,yao_2018_mvsnet,wang2020patchmatchnet,gu_2020_cascademvsnet}.

EdgeStereo~\cite{song2018edgestereo} uses a pre-trained sub-network for detecting the edges, and the edge cues are then fed into the disparity branch to improve the disparity map.  
Tosi et al.~\cite{Tosi2021CVPR} showed that it is possible to improve the quality of the learning-based two-view stereo networks by integrating an MLP-based bimodal mixture density network. 
In their work, they improved the accuracy of stereo matching networks ~\cite{chang2018pyramid,yang2019hierarchical} that were used as a backbone to their mixture density head. Inspired by these works, we also represent depth as bimodal distribution, and we jointly estimate depth maps and object boundary maps in the multi-view stereo setting using a novel multi-task learning architecture. Our pipeline does not involve any parallel (sub)networks and learns directly from multi-view images to estimate edge-depth pairs jointly. 

The continuous disparity network \cite{garg2020wasserstein} aims to regress the multi-modal depth by jointly estimating both probability and offset volume by minimizing a Wasserstein distance between the ground truth and the distribution estimated from the volumes. 
The offset volume aims to obtain continuous disparity estimations. 
Our method avoids regressing the offset values and instead, directly estimates bimodal distribution parameters. 

\subsection{Learning-based MVS} 

State-of-the-art learning-based MVS approaches adapt the photogrammetry-based MVS algorithms by implementing them as a set of differentiable operations defined in the feature space. 
MVSNet~\cite{yao_2018_mvsnet} introduced good quality 3D reconstruction by regularizing the cost volume that was computed using differentiable homography on feature maps of the reference and source images. 
Its network architecture is similar to the learning-based two-view stereo matching architecture GCNet~\cite{kendall2017end}. 
Both MVSNet~\cite{yao_2018_mvsnet} and GCNet~\cite{kendall2017end} regularize cost volume using a 3D CNN-based U-Net. The cost volume itself has a very high demand for memory. To circumvent this problem, R-MVSNet uses
GRUs~\cite{yao_2019_rmvsnet} to regularize the cost volume sequentially. 
Follow-up works~\cite{gu_2020_cascademvsnet,yang_2020_cvpr}, used feature pyramids and cost volume pyramids to learn in a coarse-to-fine manner instead of constructing a cost volume at a fixed resolution. To fully avoid the construction of feature cost volume, Wang et. al.~\cite{wang2020patchmatchnet} introduced a learning-based Multi-View PMS pipeline. Variations of PMS are seen as suitable options to work with high-resolution images since both traditional and learning-based Multi-View PMS avoids the memory demands of Plane Sweep Stereo or feature cost volume regularization.

In contrast to two-view or multi-view  Plane Sweep stereo~\cite{curless1996volumetric, zach2007globally} and cost volume regularization methods~\cite{kendall2017end, yao_2018_mvsnet}, to reduce memory consumption our pipeline estimates depth maps  by fully avoiding the cost volume creation and usage of 3D CNN networks. 
For this, we are leveraging differentiable PatchMatch-based Multi-View Stereo as part of the internal structure of our pipeline~\cite{galliani_2015_gipuma, wang2020patchmatchnet}. 

The recent work of PatchMatchNet~\cite{wang2020patchmatchnet} showed state-of-the-art results in terms of reconstruction completeness, which is used as a baseline in this work.

To enhance the quality of scene reconstruction, our proposed method focuses on estimating the geometric boundaries of objects in the scene where depth discontinuities occur. We introduce a technique to regularize the depth map by incorporating an estimated boundary map. Our approach distinguishes itself from DEF-MVSNet~\cite{lin2021high} in terms of how edge information is represented and modeled. While DEF-MVSNet primarily focuses on determining flow directions as pixel offsets, our method explicitly learns and smooths the edge map by defining each pixel as a bimodal distribution. This distinction contributes to the unique characteristics of our approach.

Similarly, our method deviates from BDE-MVSNet~\cite{ding2022adaptive}, which also aims to find flow directions for edge pixels using gradient information. Instead, we explicitly learn the boundary map, placing emphasis on regularizing smoothness in regions that are not classified as boundaries. In comparison to ElasticMVS~\cite{zhang2022elasticmvs} which proposes an elastic part representation for encoding physically connected part segmentations, our approach focuses solely on explicitly learning the boundary map. By utilizing the boundary map for regularization, our objective is to enhance smoothness rather than encode physically connected part segmentations and capture surface connectedness and boundaries within the image.

During the development of our method, we also explored some depth derivative-based loss functions, similar to those utilized in previous works~\cite{song2018edgestereo,zhang2023mg}. However, we did not observe significant improvements when employing these loss functions. Therefore, we adopted a different approach by explicitly learning the boundary map to regulate smoothness in regions that are not classified as boundaries.

In comparison to two-view stereo matching pipeline SMD-Nets~\cite{Tosi2021CVPR}, we employ a mixture density network as an internal structure for depth refinement, inputting it with RGB-Depth pairs instead of rectified left-right image pairs. Unlike previous methods, we learn the depth and boundary map simultaneously, utilizing the same backbone architecture for estimating the density parameters and boundary map in parallel. In comparison with previous methods,  our pipeline utilizes a 2D CNN-based U-Net architecture~\cite{ronneberger2015u} to estimate the bimodal depth density parameters for each pixel in discrete space.

%% file: source/method.tex
\section{Method}
\label{sec:method}

In contrast to existing MVS approaches with depth map representations, in which the depth of each pixel is expressed as a single value, our approach takes advantage of a bimodal depth representation that represents depth as distribution.
Our depth map is thus not a common grid of per-pixel scalars, but per-pixel mixture density parameters. The motivation of this module is to implicitly integrate the uncertainty notion into our pipeline, which enables us to learn depth discontinuities for spatial regularization of the depth map and to further alleviate the noise gathered in intra-object transitions, foreground-background transitions, and partial occlusions. 

\begin{figure*}[!h]
	\begin{center}
		\includegraphics[width=0.98\linewidth]{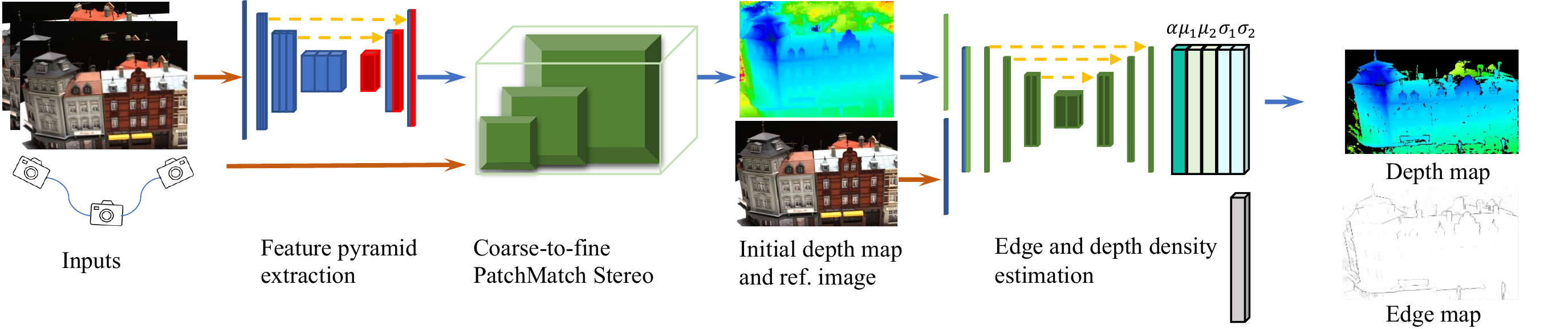}
		\caption{An overview of the proposed multi-view depth discontinuity learning network that outputs depth and edge information for each pixel.
			The brown arrows represent input feed and the blue arrows represent pipeline flow. We first extract multi-scale features from color images with FPN~\cite{lin2017feature} alike auto-encoder. Then we feed extracted features and camera parameters to the coarse-to-fine PMS module to extract the initial depth map. Using the initial depth map and RGB pair, our network learns bimodal depth parameters and geometric edge maps. We use mixture parameters and photo-geometric filtering to compute our final depth map. The edge map visualized here is negated edge map (for a clear view).}
		\label{fig:pipeline}
	\end{center}
\end{figure*}

The overview of our proposed network architecture is shown in Fig.~\ref{fig:pipeline}. 
Our network has three parts, namely, feature extraction, coarse-to-fine PatchMatch Stereo (PMS), and depth discontinuity learning, detailed as follows.

\subsection{Feature extraction}
We employ a widely used technique in the Computer Vision field known as feature pyramid learning~\cite{cheng_2020_ucsnet,gu_2020_cascademvsnet,yang_2020_cvpr}, which enables us to build our algorithm in a coarse-to-fine regression fashion. 
We adopted the Feature Pyramid Networks~\cite{lin2017feature} with residual connections between encoder and decoder, and use three layers of decoder outputs as our extracted features. 
Each subsequent level has half the resolution of the level before it, and the finest level has half the width and height of the original image. 
In Fig.~\ref{fig:pipeline}, the red blocks show three scales of features fed to the coarse-to-fine PMS module.

\subsection{Coarse-to-fine PMS}

Being agnostic to the backbone our method is independent of the underlying rough depth estimation method. 
Both cost volume regularization and the PatchMatch-based approach can be used for depth estimation. 
We follow PatchmatchNet~\cite{wang2020patchmatchnet} that demonstrates good reconstruction completeness and low memory demands. Our pipeline regress three levels of initial depth maps in a coarse-to-fine manner.

We randomly initialize the depth values at the coarsest level, and at a finer level, we initialize the depth values with the outputs of the coarser levels.
Following the initialization step, we run an iterative feedback loop between the propagation and evaluation steps. 
We propagate our estimates with good scoring values to the neighboring pixels.
In the evaluation step, we use candidate depth values for differentiable homography warping and matching cost computation. 

\subsection{Depth discontinuity learning}
The output of coarse-to-fine PMS is a conventional depth map of half the resolution (half width and half height) of the original input.
Hui et al.~\cite{hui16msgnet} showed that a low-resolution depth map can be progressively upsampled with the guidance of the associated high-resolution color image. 
This idea inspires the proposed framework's attempt to match the resolution of the color and depth images.

Contrary to existing learning-based networks~\cite{yao_2018_mvsnet,yu2021attention}, which revise depth maps using residual networks, we refine depth maps via learning mixture density parameters and geometric edge maps. 
In contrast to SMD-Nets~\cite{Tosi2021CVPR}, which employ corrected image pairs as input, we use RGB-depth pairs as the input to the depth refinement network and convolutional mixed density networks as the internal structure.
To the best of our knowledge, our work is the first learning-based MVS method that explicitly learns depth discontinuity maps (aka geometric edge maps) to simultaneously refine the quality and improve the smoothness of the depth maps.

In our pipeline, we use a 2D CNN-based U-Net~\cite{ronneberger2015u} architecture to estimate the bimodal depth density parameters of each pixel in a discrete space. 
During the development of our pipeline, we experimented with different network variations to learn separate boundary maps and mixture density parameters, including two parallel network streams and single encoder and multiple decoder architectures. However, we found that using multiple subnetworks increased the number of parameters and GPU memory demands without leading to any substantial improvement in results. Therefore, we chose to use  a single encoder and decoder architecture for our proposed pipeline.

Based on the fact that depth maps have piecewise smoothness and that they can be improved by spatial regularization to smooth regions as shown in earlier works~\cite{huang00,boykov2001fast,boykov_cvpr_1998}, we propose to refine depth-map quality by learning depth discontinuities. 

Previous methods based on pixel-wise single value estimates implicitly balance the depth estimation error between nearby foreground and background pixels for boundary points. 
Our refinement network regresses the parameters of a  bimodal distribution.
We use the bimodal Laplacian distribution, which was inspired
by Tosi et al.~\cite{Tosi2021CVPR} work. 
During development, we observed that the Laplacian distribution~\cite{laplace_distribution} had slightly better results than the Gaussian. The Laplacian distribution has a sharper shape modality than Gaussian. It optimizes over $\mathcal{L}_1$ distance instead of $\mathcal{L}_2$ distance between the groundtruth and estimated mean. This makes it more robust against outliers. The bimodal Laplacian density distribution can be written as 

\begin{equation}
\label{eq:laplacian}
\begin{gathered}
\theta = \{ \alpha,\mu_1,\sigma_1,\mu_2,\sigma_2\} \\ 
\end{gathered}\\
\end{equation}
\begin{equation}
\begin{gathered}
p(x;\theta)=
\frac{ \alpha}{2\sigma_1}\exp(-\frac{|x-\mu_1|}{\sigma_1}) + \frac{1- \alpha}{2\sigma_2}\exp(-\frac{|x-\mu_2|}{\sigma_2})
\end{gathered}
\nonumber
\end{equation}
where $ \alpha$ is the mixture weight that can be seen as the likeliness of each mode. 
Later in our work (see Sec.~\ref{sec:exp}), we observe that the network learns to assign different $ \alpha$ values to different scene parts, and in most cases it is binary classifying foreground and background pixels. 
$\mu_1$ and $\mu_2$ are the two depth estimates of the corresponding modes. 
$\sigma_1$ and $\sigma_2$ are the two depth variance measures of each depth value. We also treat $\frac{ \alpha}{\sigma_1}$ and $\frac{1- \alpha}{\sigma_2}$ as responsibility scores, which aims to determine the responsible mode for the depth of a given pixel. 

Besides extending bimodal depth estimation to the multi-view case, our proposed convolutional mixture density network also shows that with a single stream compact discontinuity learning network architecture, it is possible to achieve three goals: (1) Upsampling; (2) Refining; (3) Multi-task learning.

\subsection{Loss function}
Our loss function has four terms: Depth-groundtruth loss, Edge-depth loss, Smoothness loss, and Bimodal depth loss, each defined with a specific purpose.

\textbf{Depth-groundtruth loss}.
This loss term measures the difference in depth maps between prediction and the groundtruth. 
It is defined as the mean absolute error (MAE) of the estimated depth map, i.e., $\mathcal{L}_1$ distance between the estimated depth and ground-truth depth across all stages of the PMS and the final reconstructed depth,
\begin{equation}
\begin{gathered}
L_{gt}  = \sum_{k=0}^{3}[\frac{1}{N_k} \mathcal{L} _1(D_k,\hat{D_k})],
\end{gathered} 
\label{eq:DepthGT}
\end{equation}
where $k \in \{0, 1, 2, 3\}$ denotes the scale index of the coarse-to-fine PMS that estimates initial low-resolution depth maps, with 0 representing the finest input and output resolution, and from 3 to 1 the coarser-to-finer scales of the PMS output.
$\hat{D_k}$ and $D_k$ represent the ground-truth depth map and estimated depth map at resolution level $k$, respectively.  The DTU dataset~\cite{aanaes2016_dtu} contains masks that identify pixels with valid ground truth depth information. $N_k$ represents the number of pixels in each scale.

\textbf{Edge-depth loss}.
Geometric edges or boundaries are expected where there are depth discontinuities in the depth map. 
Thus, the edge-depth loss term measures how much the estimated edges agree with the second-order depth variations (i.e., depth discontinuities).  
It is defined as the mean squared error (MSE) ($\mathcal{L}_2$ distance) between the estimated edge $E$ and groundtruth changes of variations in depth $\hat{D}$,
\begin{equation}
\begin{gathered}
L_{ed}  = \frac{1}{N} \mathcal{L} _2(E,\phi(\Delta\hat{D},\tau)),
\end{gathered} 
\label{eq:EDGE-DEPTH}
\end{equation}
where $\phi$ is the function that takes Laplacian of the depth and threshold value $\tau$ to return the mask image where the Laplacian response~\cite{laplace_operator} of the depth map is higher than the $\tau$.  The DTU dataset~\cite{aanaes2016_dtu} contains masks that identify pixels with valid ground truth depth information. The variable $N$ represents the count of masked pixels that have corresponding ground truth labels for depth.
With this term, we explicitly inform the network that we are expecting geometric edges or boundaries at the pixels where there exist depth discontinuities.
We calculate depth discontinuities using the Laplacian operator, which is the second-order depth change.

\textbf{Smoothness loss}.
Except for the geometric edges and boundaries with depth discontinuities, real-world objects typically demonstrate piecewise smoothing surfaces. Thus, we would like to encourage local smoothness for the regions without depth discontinuities. We achieve this by introducing an edge-aware smoothness loss term to penalize second-order depth variations in non-boundary regions,
\begin{equation}
\begin{gathered}
L_{sm}  = \frac{1}{N} \sum_{i \in \Omega}^{}\omega(E_i)|\Delta D_i|\\
\omega(E_i)  = \exp(-\beta E_i)\\
\end{gathered},
\label{eq:Smooth}
\end{equation}
where $E_i$ will have an estimated value close to 1 for boundaries and close to 0 for non-boundary pixels. $\omega$ is a weight function that plays a role of a switch, which returns a value close to 0 for boundaries and close to 1 for non-boundary pixels. Thus, second-order depth change in non-boundary regions contributes to our smoothness loss. $\beta$ is a tunable hyper-parameter that controls the sharpness of change in the $\omega$ function. $N$ denotes the number of pixels in the image space $\Omega$ with a valid grountruth depth.
To the best of our knowledge, this is the first time depth discontinuities are explicitly learned and used for spatial regularization in multi-view stereo networks.

\textbf{Bimodal loss}.
We adopt a common approach of minimizing the negative-log likelihood of the distribution to increase the likelihood of true depth. Tosi et al.~\cite{Tosi2021CVPR} have demonstrated that this loss term for bimodal depth in the two-view stereo setting can produce inspiring results. Our bimodal loss term is defined as

\begin{equation}
\begin{gathered}
L_{bi}  =  \frac{1}{N} \sum_{i \in \Omega}^{}- \log(p(\hat{D_i};\theta,i)),
\end{gathered} 
\label{eq:BimodalLOSS}
\end{equation}
where $\hat{D_i}$ represents the groundtruth depth measured at pixel $i$, and $\theta$ is the parameter of the bimodal distribution introduced in Eq.~\ref{eq:laplacian}. The distribution $p$ can be computed using the Eq.~\ref{eq:laplacian}. $N$ denotes the number of the pixels in the image space $\Omega$ with a valid grountruth depth.

\textbf{Total loss}.
We simply use the weighted sum of the aforementioned loss terms
\begin{equation}
\begin{gathered}
L_{total}  =  L_{gt} + \lambda_1 L_{ed} + 
\lambda_2 L_{sm} + \lambda_3 L_{bi}
\end{gathered}
\label{eq:TOTALLOSS}
\end{equation}
as a training criterion for our network to optimize the parameters via backpropagation.
 $\lambda_1 =4$, $\lambda_2 =1.25$, and $\lambda_3 =0.5$ are hyper-parameters empirically set based on our experiments on the validation set.

%% file: source/experiments.tex
\section{Experiments and Evaluation}
\label{sec:exp}
We used the same model to quantitatively evaluate the generalization capabilities of our method and to compare it with other methods. All the metric results of the other methods were collected from the corresponding papers, and the 3D point clouds of other papers were reconstructed using the code and pre-trained models provided by the authors. 
\subsection{Datasets}

We have tested and evaluated our method on multiple datasets: the small baseline dataset DTU~\cite{aanaes2016_dtu}, and the large baseline datasets ``Tanks and Temples''~\cite{Knapitsch2017}, ETH3D~\cite{schoeps2017cvpr} and BlendedMVS~\cite{yao2020blendedmvs}. 

The DTU dataset~\cite{aanaes2016_dtu} is a benchmark with 120 scenes captured by a structured-light sensor under seven different lighting conditions. 
It has been widely used for developing learning-based MVS methods and evaluating their performance in terms of completeness and accuracy. 
All the learning-based methods in Tab.~\ref{tab:dtu_eval} are trained on the same 79 scans, validated on the same 18 scans, and evaluated on the remaining 22 scans.
\footnote{Validation set: scans ${3, 5, 17, 21, 28, 35, 37, 38, 40, 43, 56, 59, 66, 67, 82, 86, 106, 117}$. \\
	Evaluation set: scans ${1, 4, 9, 10, 11, 12, 13, 15, 23, 24, 29, 32, 33, 34, 48, 49,62, 75, 77, 110, 114, 118}$.}

``Tanks and Temples" is a real-world large-scale dataset consisting of both indoor and outdoor scenes~\cite{Knapitsch2017}. It has two parts: an intermediate set consisting of images of sculptures, large vehicles, and house-scale buildings (taken from the exterior), and an advanced set consisting of images of large indoor scenes and large outdoor scenes with complex geometric layouts and repetitive structures. 

The ETH3D dataset~\cite{schoeps2017cvpr} is a collection of calibrated images, containing various indoor and outdoor environments, including urban scenes, garages, and rooms. The dataset provides ground truth camera poses, 3D point cloud geometry, and images for each scene, making it suitable for tasks such as camera pose estimation and 3D reconstruction.

BlendedMVS~\cite{yao2020blendedmvs} is a large-scale MVS dataset for generalized multi-view stereo networks. The dataset contains samples covering a variety of scenes, including architecture, sculptures, aerial images, and small objects.

\begin{figure*}[t]
	\centering
	\includegraphics[width=0.9\linewidth]{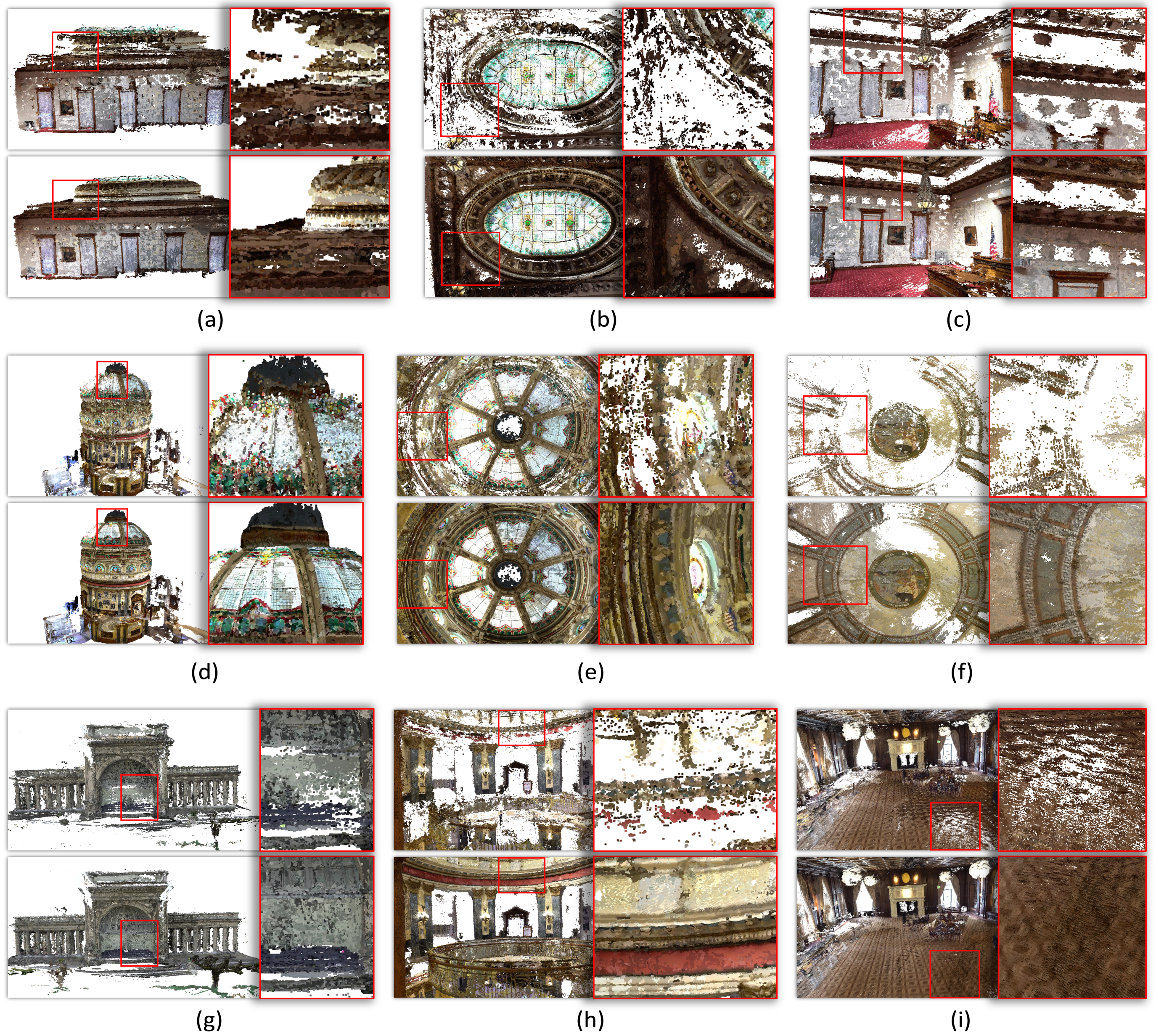}
	\caption{Comparison between our method and the baseline method PatchmatchNet~\cite{wang2020patchmatchnet} on a set of scenes from the Tank and Temples dataset~\cite{Knapitsch2017}. For each scene, the top row shows the results from PatchmatchNet, and the bottom row shows the results from our method. A zoomed view of the marked image region is shown on the right of each result.}
	\label{fig:TNT_comparison}
\end{figure*}

\subsection{Evaluation on DTU dataset}

In this section, we present our findings based on the DTU benchmark~\cite{aanaes2016_dtu}, where we evaluated the performance of our method using the \textit{accuracy}, \textit{completeness}, and \textit{overall} metrics. The \textit{accuracy} metric measures the mean error distance between the closest points in the reconstruction and the reference based on structured light. The \textit{completeness} metric quantifies the mean error distance between the closest points in the reference and the reconstruction. The \textit{overall} metric is the algebraic mean of \textit{accuracy} and \textit{completeness}. Lower scores indicate better performance in this benchmark.

The result on the DTU dataset is reported in Tab.~\ref{tab:dtu_eval}. For a fair comparison, all techniques were trained on the same dataset and employed the same validation and train split.
From the result, we can see that traditional photogrammetry-based methods generally have better accuracy, while learning-based methods have better completeness and \textit{overall} performance. 
Furthermore, it also reveals that the \textit{completeness} gap between learning-based and photogrammetry-based methods is bigger than their gap in \textit{accuracy}, which motivated us to use a coarse-to-fine PMS to build our initial depth estimation block, to reduce the \textit{accuracy} gap while still improving completeness.

During a similar development phase, several methods such as UniMVSNet ~\cite{peng2022rethinking} and TransMVSNet~\cite{wang2022mvster} have been published, showcasing superior performance compared to our proposed approach. However, despite these advancements, we maintain a strong belief in the effectiveness of our Depth Discontinuity Learning (DDL) module in enhancing the baseline performance of PatchmatchNet.

This reveals that learning depth discontinuities is an effective means to improve both reconstruction accuracy and completeness.

\begin{table}[t]
	\centering
	\begin{widetable}{0.1\textwidth}{ lccc}
		\hline
		Method & \begin{tabular}[c]{@{}c@{}}Accuracy\\($mm$) $\downarrow$ \end{tabular} & \begin{tabular}[c]{@{}c@{}}Completeness\\($mm$) $\downarrow$ \end{tabular} & \begin{tabular}[c]{@{}c@{}}Overall\\($mm$) $\downarrow$ \end{tabular}\\
		\hline
		\multicolumn{4}{c}{Traditional photogrammetry-based} \\
		\hline
		Camp~\cite{camp} & 0.835 & 0.554 & 0.695 \\ 
		Furu~\cite{furukawa2010} & 0.613 & 0.941 & 0.777 \\
		Tola~\cite{tola2012efficient} & 0.342 & 1.190 & 0.766 \\
		Gipuma~\cite{galliani_2015_gipuma} & \textbf{0.283} & 0.873 & 0.578 \\
		\hline
		\multicolumn{4}{c}{Learning-based} \\
		\hline
		SurfaceNet~\cite{ji_2017_surfacenet}  & 0.450 & 1.040 & 0.745 \\
		MVSNet~\cite{yao_2018_mvsnet} & 0.396 & 0.527 & 0.462\\
		R-MVSNet~\cite{yao_2019_rmvsnet} & 0.383 & 0.452 & 0.417\\
		CIDER~\cite{xu2020learning_inverse} & 0.417 & 0.437 & 0.427\\
		P-MVSNet~\cite{luo_2019_p_mvsnet} & 0.406 & 0.434 & 0.420\\
		Point-MVSNet~\cite{chen_2019_pointmvsnet} & 0.342 & 0.411 & 0.376\\
		AttMVS ~\cite{Luo_2020_CVPR} & 0.383 & 0.329 & 0.356\\
		Fast-MVSNet~\cite{Yu_2020_fastmvsnet} & 0.336 & 0.403 & 0.370\\ 
		Vis-MVSNet~\cite{zhang2020visibility} & 0.369 & 0.361 & 0.365\\
		CasMVSNet~\cite{gu_2020_cascademvsnet} & 0.325 & 0.385 & 0.355\\
		UCS-Net~\cite{cheng_2020_ucsnet} & 0.338 & 0.349 & 0.344 \\
		EPP-MVSNet~\cite{ma2021epp} & 0.413 & 0.296 & 0.355 \\
		CVP-MVSNet~\cite{yang_2020_cvpr} & 0.296 & 0.406 & 0.351 \\
		AA-RMVSNet~\cite{wei2021aa} & 0.376 & 0.339 & 0.357 \\
		DEF-MVSNET~\cite{lin2021high} & 0.402 & 0.375 & 0.388 \\
		ElasticMVS~\cite{zhang2022elasticmvs}   & 0.374 &  0.325 & 0.349\\
		MG-MVSNET~\cite{zhang2023mg} & 0.358 & 0.338 & 0.348 \\
		BDE-MVSNet~\cite{ding2022adaptive} & 0.338 & 0.302 & 0.320 \\
		UniMVSNet~\cite{peng2022rethinking} &0.352 &0.278 &0.315\\
		TransMVSNet~\cite{wang2022mvster} & 0.321 &  0.289 & \textbf{ 0.305}\\          	
		\hline
		PatchmatchNet ~\cite{wang2020patchmatchnet} & 0.427 & 0.277 & 0.352\\
		PatchmatchNet + Ours ($L_{1,4}$) & 0.405 & \textbf{0.267} & 0.336\\
		PatchmatchNet + Ours ($L_{1,2,3,4}$)&  0.399  & 0.280 & 0.339\\
		\hline	
	\end{widetable}
	\caption{Quantitative comparison with photogrammetry-based and learning-based MVS methods, on the DTU dataset~\cite{aanaes2016_dtu}. Two different settings (with different loss functions) of our method were tested. $L_{1}$: depth-groundtruth loss; $L_2$: edge-depth loss; $L_3$: smoothness loss; $L_{4}$: bimodal loss. Please note that the metrics are error-based and thus the smaller the better. }
	\label{tab:dtu_eval}
\end{table}

\begin{table}[t]
	\centering
	\begin{widetable}{\columnwidth}{cccc}
		\hline
		Method & \begin{tabular}[c]{@{}c@{}}Accuracy\\($\%$) $\uparrow$ \end{tabular} & \begin{tabular}[c]{@{}c@{}}Completeness\\($\%$) $\uparrow$ \end{tabular} & \begin{tabular}[c]{@{}c@{}}F-score $\uparrow$ \\\end{tabular}\\
		\hline
		PatchmatchNet & 64.81 & \textbf{65.43} & 64.21\\
		Ours & \textbf{64.96} & 65.21& \textbf{64.37}\\
		\hline	
	\end{widetable}
	\caption{Quantitative evaluation of our method and comparison with PatchmatchNet~\cite{wang2020patchmatchnet} on the ETH3D training set~\cite{schoeps2017cvpr}. Following the benchmark, the \textit{accuracy} and \textit{completeness} measures are quantified using the percentage of points below a 2 $cm$ error margin (the higher the better).}
	\label{tab:eth3_eval}
\end{table}

\begin{table*}[t]
	\begin{widetable}{\textwidth}{c|ccc|ccc}
		\hline
		&    & Intermediate set & &  & Advanced set & \\
		\hline
		Methods & P (\%) $\uparrow$ & R (\%) $\uparrow$ & F-score $\uparrow$ & P (\%) $\uparrow$ & R (\%) $\uparrow$ & F-score $\uparrow$ \\
		\hline
		
		PatchmatchNet & 43.64 &  69.38 & 53.15 & 27.27 & \textbf{41.66 }& 32.31\\
		Ours  & \textbf{45.12} & \textbf{69.69} & \textbf{54.30}& \textbf{28.31} & 41.06 & \textbf{32.80}\\
		\hline
	\end{widetable}
	\caption{Evaluation and comparison with PatchmatchNet~\cite{wang2020patchmatchnet} on the ``Tanks and Temples" dataset~\cite{Knapitsch2017}.}
	\label{tab:TNT_eval}
\end{table*}

\begin{table*}[!t]
	\begin{widetable}{\textwidth}{c|ccc|cc}
		\hline
		&  \multicolumn{3}{|c|}{Point clouds (testing)} & \multicolumn{2}{c }{Depth maps (validation)} \\
		\hline
		Methods & \begin{tabular}[c]{@{}c@{}}Acc.\\($mm$) $\downarrow$ \end{tabular} & \begin{tabular}[c]{@{}c@{}}Comp.\\($mm$) $\downarrow$ \end{tabular} & \begin{tabular}[c]{@{}c@{}}Overall\\($mm$) $\downarrow$ \end{tabular} & \begin{tabular}[c]{@{}c@{}}Depth map \\($mm$) $\downarrow$ \end{tabular} & \begin{tabular}[c]{@{}c@{}}Error ratio\\(\%; error $>$  8 $mm$) $\downarrow$ \end{tabular} \\
		\hline
		PatchmatchNet ~\cite{wang2020patchmatchnet} & 0.427 & 0.277 & 0.352 & 7.09 & 11.58\\
		Architecture + $L_{1}$  & 0.412 & 0.273 & 0.342  & 5.41 & 9.07 \\
		Architecture + $L_{1,2,3}$  & 0.412 &  0.270  & 0.341   & 5.44  &  8.96 \\
		Architecture + $L_{1,4}$  & 0.405 & \textbf{0.267} & \textbf{0.336}   &5.47 &  9.01\\
		Architecture + $L_{1,2,3,4}$ &\textbf{ 0.399} & 0.280 &  0.339  &\textbf{ 5.28} & \textbf{8.79} \\
		\hline
	\end{widetable} 
	\caption{Ablation study on the point clouds and depth maps from the DTU dataset~\cite{aanaes2016_dtu}. $L_{1}$: depth-groundtruth loss; $L_2$: edge-depth loss; $L_3$: smoothness loss; $L_{4}$: bimodal loss. Note that $L_2$ and $L_3$ cannot be separated because they together work for edge-aware smoothness.}
	\label{tab:ablation_dtutest_eval}
\end{table*}

\subsection{Evaluation on ``Tanks and Temples'' dataset}

In this section, we present our findings based on the `` Tanks and Temples'' dataset~\cite{Knapitsch2017}.
This benchmark has three metrics, namely, recall, precision, and F-score. Recall and precision represent the completeness and accuracy of the reconstruction, respectively, both measured in percentage (\%). The F-score combines precision and recall, and it is defined as the harmonic mean of a model’s precision and recall.

In our experiments, we used our model trained using the DTU dataset with 14 epochs with all the proposed loss terms. 
We compared the results against those from our baseline method PatchmatchNet~\cite{wang2020patchmatchnet}. 
For both methods, we ran the same depth map fusion algorithm with the same threshold value to not gain any advantage in the evaluation process. 
As can be seen from the statistics reported in Tab.~\ref{tab:TNT_eval}, our results on the intermediate set have better performance on all evaluation metrics.
On the advanced set, our results demonstrate better accuracy and F-score, and the results from PatchmatchNet have slightly better completeness.
As depicted in Fig.~\ref{fig:TNT_comparison}, our approach improves baseline ~\cite{wang2020patchmatchnet} in accurately capturing the overall geometry and exhibits improved completeness in smooth regions. This is substantiated by both qualitative and quantitative results, which demonstrate that our approach outperforms the baseline in terms of overall reconstruction quality.

\subsection{Evaluation on ETH3D dataset}
In this section, we present our findings based on the ETH3D benchmark~\cite{schoeps2017cvpr}. 
The ETH3D benchmark~\cite{schoeps2017cvpr} consists of high-resolution images of scenes with sparse scene coverage, high viewpoint variation, and camera parameter information. The quantitative evaluation of our method and the comparison with PatchmatchNet~\cite{wang2020patchmatchnet} on the ETH3D dataset~\cite{schoeps2017cvpr} are detailed in Tab.~\ref{tab:eth3_eval}. Both methods have used the same fusion pipeline. Our method demonstrates better accuracy and F-score, while PatchmatchNet has better completeness.

\subsection{Ablation study}

We have conducted an ablation study to understand and analyze the contributions of the aforementioned loss terms of our architecture. The results are detailed in Tab.~\ref{tab:ablation_dtutest_eval}. 
Since the edge-depth loss and the smoothness loss terms together strive for edge-aware smoothness, we do not separate them in our experiments.
We retrieve the last two metrics from the validation set while tuning our hyper-parameters. The ``Depth map'' represents the accuracy of the estimated depth map, calculated using mean absolute error (MAE) between the estimated depth map and groundtruth. ``Error $>$ 8 $mm$'' represents the percentage of points in the depth map having a higher error than 8 $mm$. 

From Tab.~\ref{tab:ablation_dtutest_eval}, we can see that using all lost terms improves the depth map quality on the validation set.
For testing, we observe that our point clouds have better completeness and overall metrics with bimodal and depth ground-truth loss while having edge-aware smoothness term results in better accuracy. Our network also improves the arithmetic mean of accuracy and completeness if we compare it against the baseline.

\begin{figure*}[!h]
	\centering
	\subfloat[Edge maps]{\includegraphics[width=0.44\linewidth]{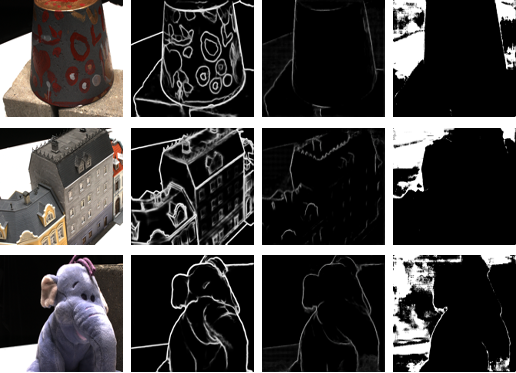}\label{subfig:edge_maps}}
	\hspace{1em}
	\subfloat[GPU memory consumption]{\includegraphics[width=0.46\linewidth]{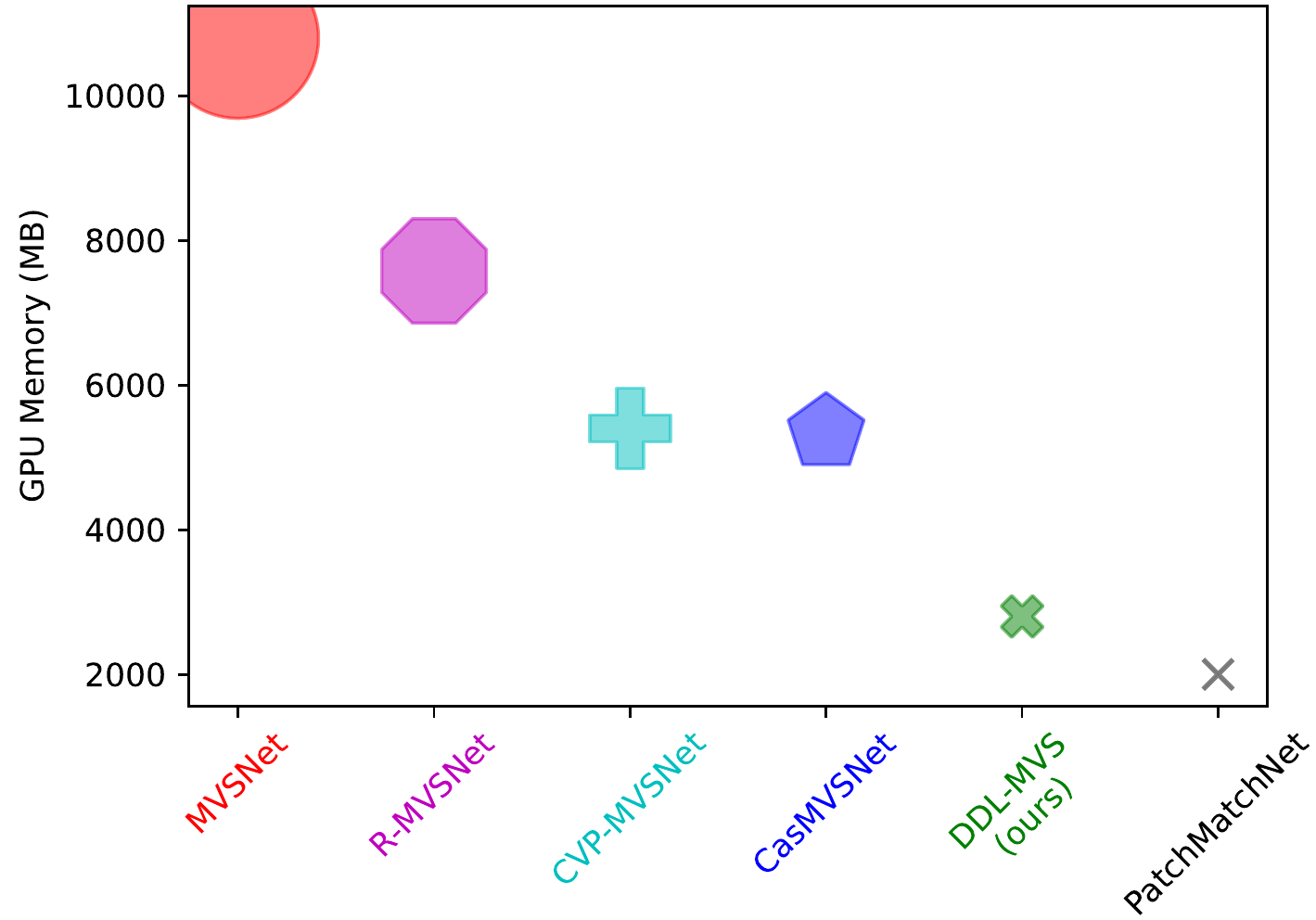}\label{subfig:memory_demand}}
	\caption{Edges maps and GPU Memory consumption. (a) Edges maps of a few randomly chosen examples. For each example, the images from left to right are the color image, the edge map predicted by HED~\cite{xie2015holistically}, our learned edge map, and the $\alpha$ map, respectively. We can see that our learned edge maps better capture the depth discontinuities, regardless of the photometric changes.
		It is also interesting to observe that our $\alpha$ maps distinguish between foreground and background. (b) Comparison of GPU memory demands with existing learning-based MVS networks on DTU dataset with image size 1152 $\times$ 864.}
	\label{fig:edgemaps_memory}
\end{figure*}

\subsection{Effect of depth discontinuity learning} 
\begin{table}[H]
	\caption{Evaluation of  depth map errors in boundary and smooth regions using the DTU dataset~\cite{aanaes2016_dtu}.}
	\label{tab:boundary_region_dtu}
	
	\begin{adjustwidth}{-\extralength}{0cm}
		\newcolumntype{C}{>{\centering\arraybackslash}X}
		\begin{tabularx}{\fulllength}{c CCC}
			\toprule
			&  \multicolumn{2}{c}{\textbf{Boundary and Smooth Region}} & \multicolumn{1}{c }{\textbf{Depth Maps} } 
			\\
			\midrule
			\textbf{Methods} & \begin{tabular}[c]{@{}c@{}}\textbf{Boundary Region (mm)} $\downarrow$ \end{tabular} & \begin{tabular}[c]{@{}c@{}}\textbf{Smooth Region (mm)} $\downarrow$ \end{tabular} & \begin{tabular}[c]{@{}c@{}}\textbf{Whole Depth Map  (mm)} \boldmath{$\downarrow$} \end{tabular}  \\
			\midrule
			PatchmatchNet ~\cite{wang2020patchmatchnet} & 22.05 & 6.66 & 7.09 \\
			Ours & \textbf{19.86}   & \textbf{4.84}  & \textbf{5.28}  \\
			\bottomrule
		\end{tabularx}
		
	\end{adjustwidth}
	
\end{table}

\begin{figure*}[t]
	\centering
	\includegraphics[width=0.9\linewidth]{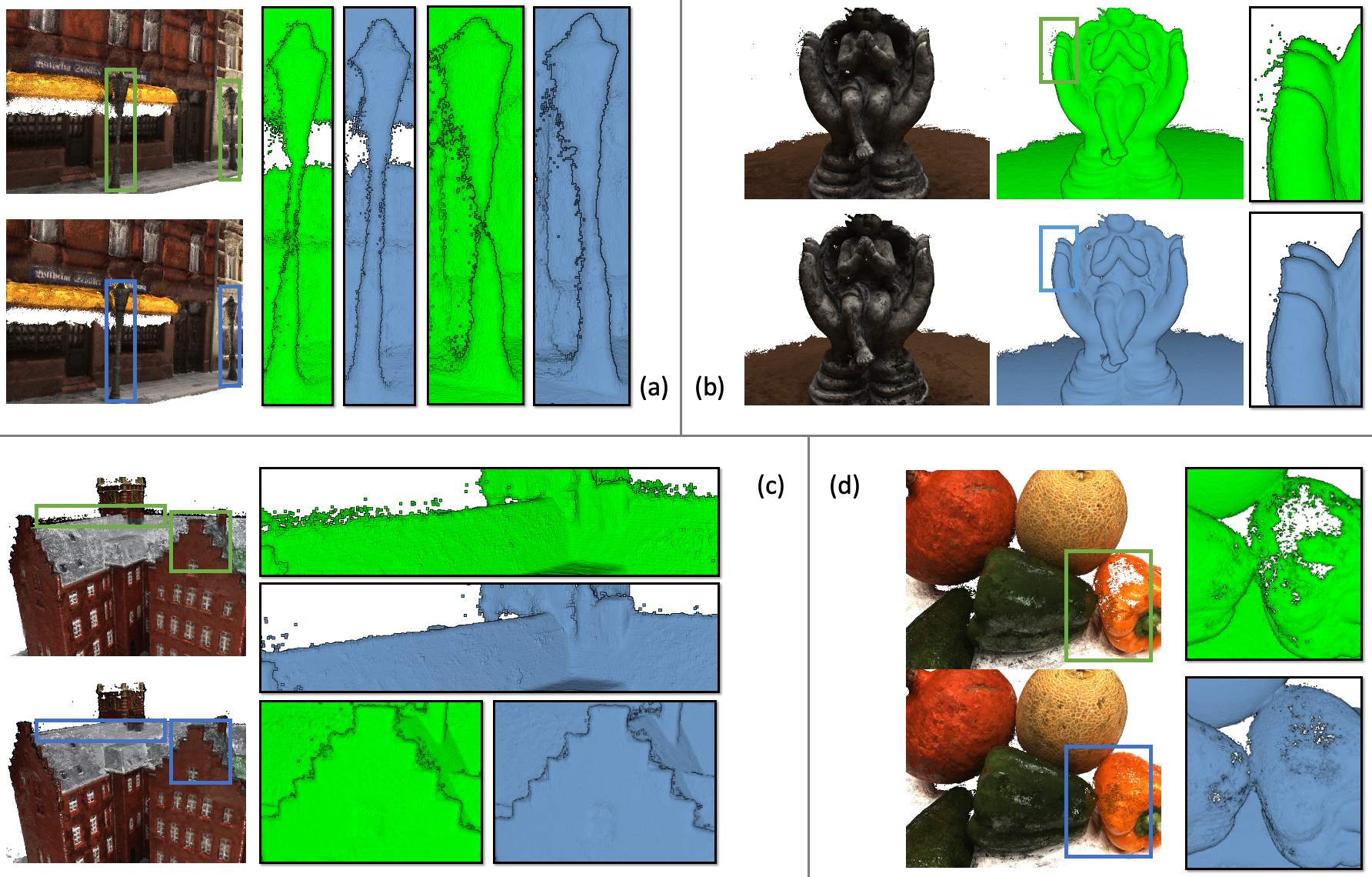}
	\caption{Comparison between our method and the baseline method PatchmatchNet~\cite{wang2020patchmatchnet} on a set of scenes from the DTU dataset~\cite{aanaes2016_dtu}. For each scene, the colored image at the top shows with green boxes the results from PatchmatchNet, and the colored image at the bottom with blue boxes shows the results from our method. A zoomed view of each marked image region is shown on the right of each result. The blue images depict our results, while the green images depict the baseline results.}
	\label{fig:DTU_boundary}
\end{figure*}

From the above experiments and evaluation, our method demonstrates superior reconstruction quality in terms of \textit{completeness} and \textit{overall} quality, which benefits from our depth discontinuity learning. 
To understand the role of depth discontinuity learning in reconstruction, we visualize the learned depth discontinuities (denoted as edge maps) for a few randomly picked examples in Fig.~\ref{fig:edgemaps_memory} (a), and compare them with the edge maps predicted using the seminal learning-based edge detection method HED~\cite{xie2015holistically}. 
We can see that by learning depth discontinuities, our network can retrieve edges where the true depth discontinuities lie. Thus, as a key component for learning-based MVS pipelines, our discontinuity-aware depth learning is more robust to photometrical changes, shadows, and small variations in depth. In the earlier stage of the development of DDL-MVS, we tried to feed the network with HED~\cite{xie2015holistically} output and jointly refine the depth and edge maps similar to EdgeStereo \cite{song2018edgestereo}. It turned out that even after refinement, the edges were too sensitive to photometric changes, leading to higher depth errors.

To reveal how our depth discontinuity learning contributes to depth estimation, we demonstrate the $\alpha$ map of each example in the last column of Fig.~\ref{fig:edgemaps_memory} (a), where $\alpha$ is the mixture weight in the bimodal Laplacian density distribution (see Eq.~\ref{eq:laplacian}). 
It is surprisingly interesting to observe that our network tries to learn to differentiate foreground and background, for which the $\alpha$ values express a binary classification for foreground and background pixels.

Our suggested framework enhances the quality of depth maps for both smooth and boundary regions, as demonstrated quantitatively in Tab.~\ref{tab:boundary_region_dtu}. We have computed mean absolute error (MAE) between the estimated depth and the groundtruth.  In contrast to the rest of the pixels, which correspond to a smooth area, boundary pixels are those pixels where the laplacian of the groundtruth depth is greater than 5.

The proposed approach also enhances the quality of the point clouds as demonstrated qualitatively in Fig.~\ref{fig:DTU_boundary}, from which we can see that thin structures and smooth regions are captured more completely, and the boundary regions have a lower amount of noise.

\begin{figure*}[!t]
	\centering
	\begin{tabular}{c@{}c@{}c@{}c@{}}
		\includegraphics[width=0.235\linewidth]{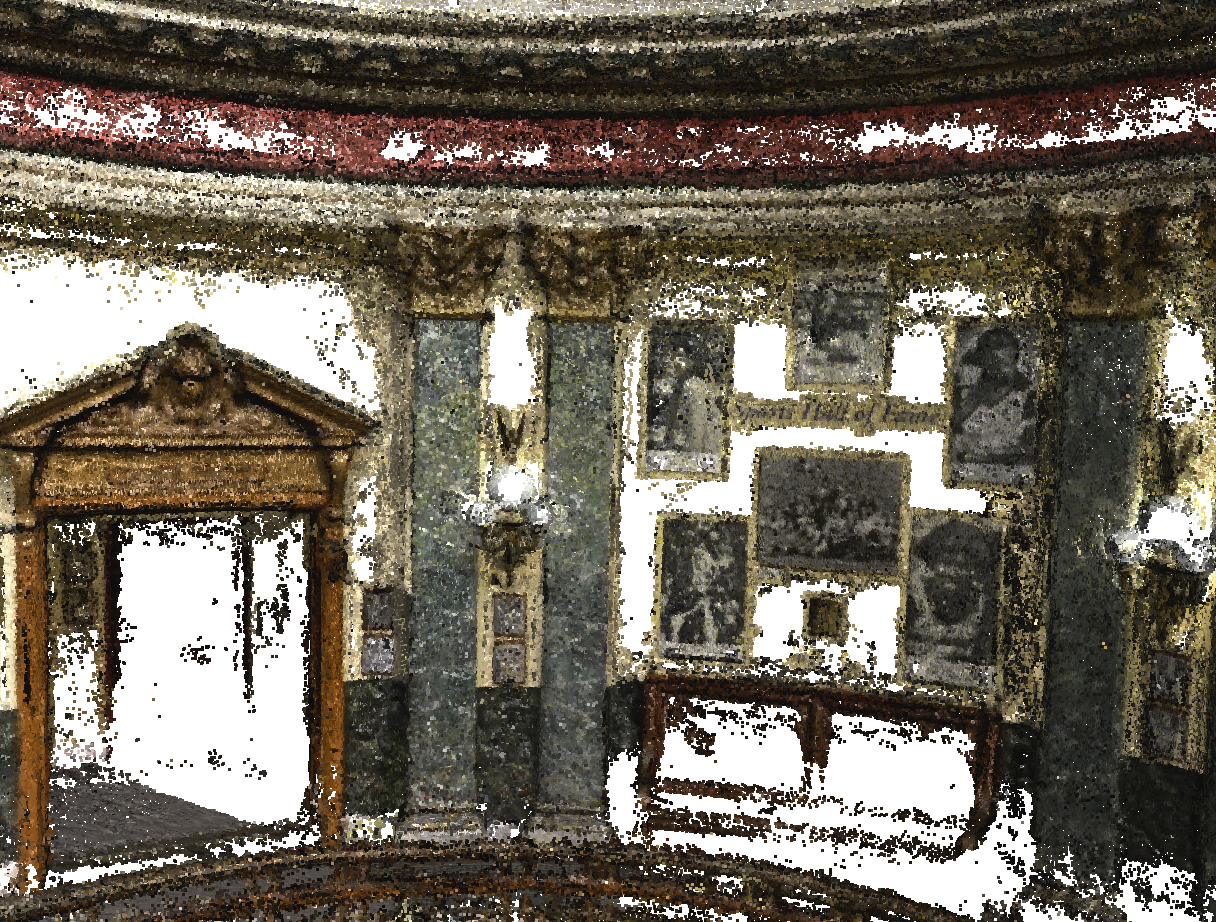} 
		\includegraphics[width=0.235\linewidth]{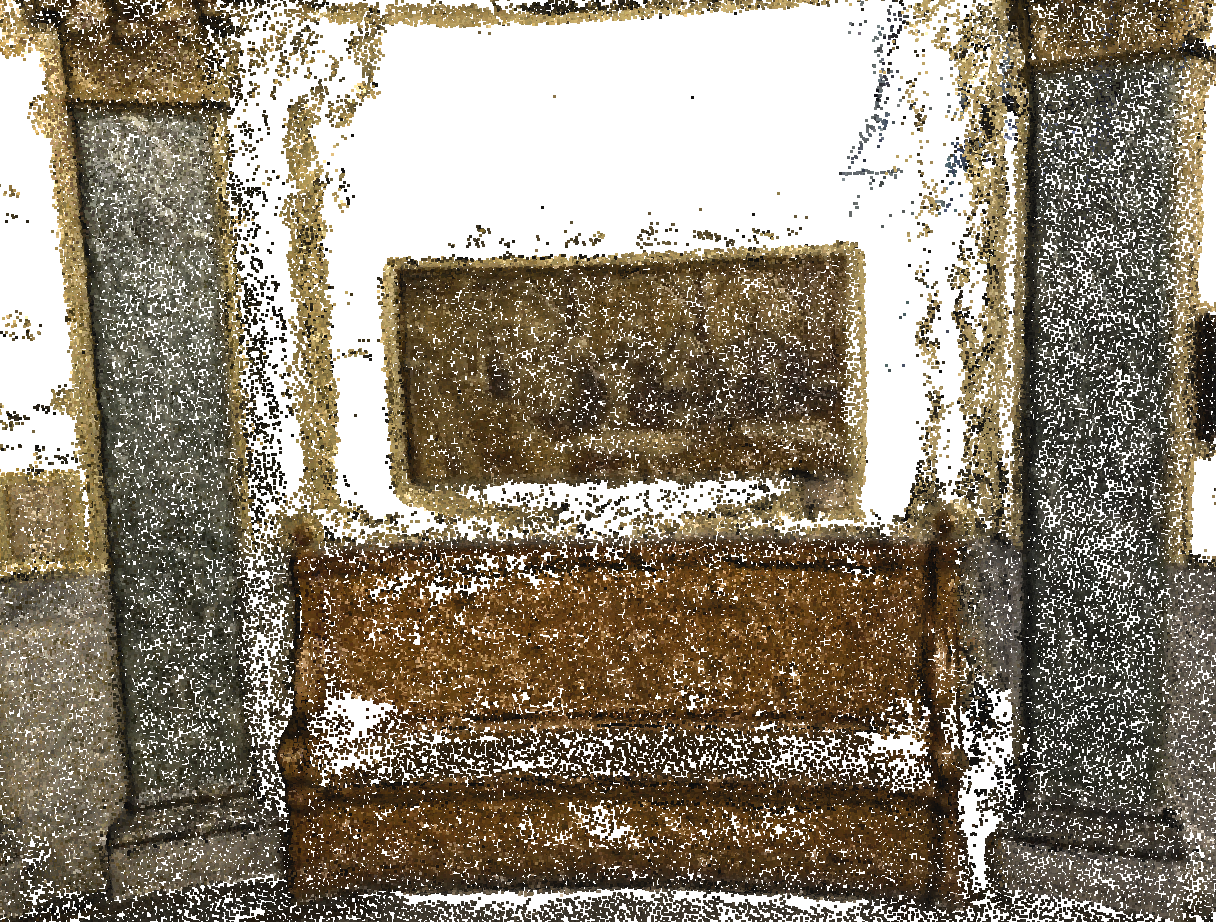} 
		\includegraphics[width=0.235\linewidth]{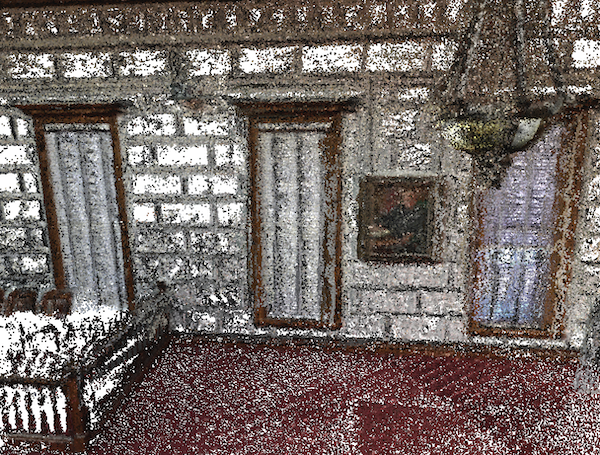} 
		\includegraphics[width=0.22\linewidth]{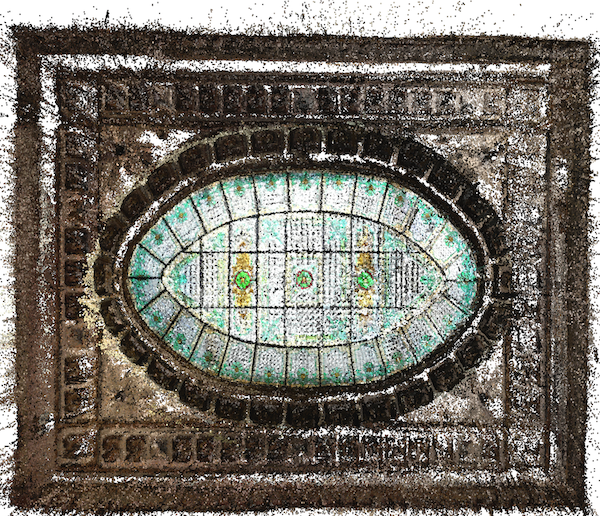}\\
		{COLMAP~\cite{schoenberger2016mvs}}
		\label{subfig:COLMAP}
	\end{tabular}
	\begin{tabular}{c@{}c@{}c@{}c@{}}	
		\includegraphics[width=0.235\linewidth]{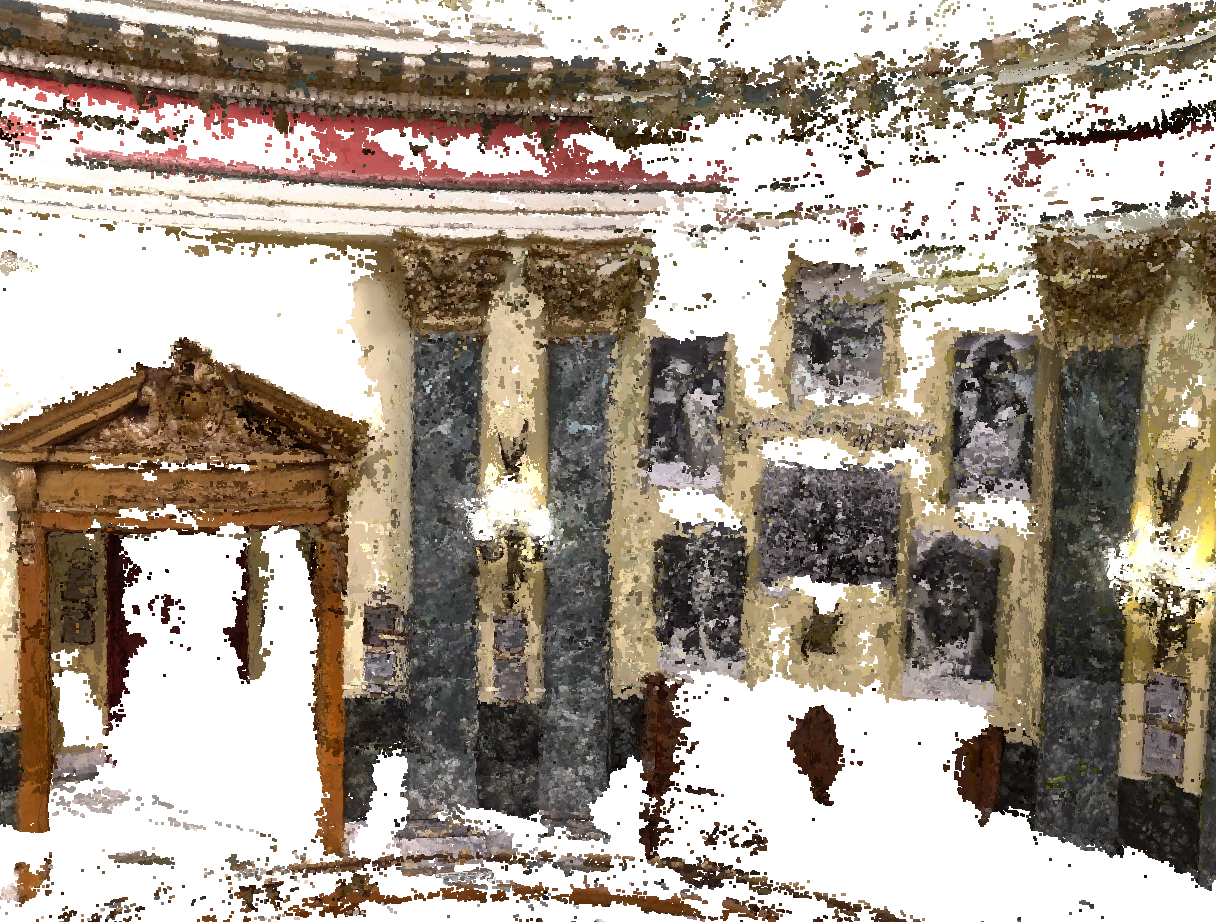} 
		\includegraphics[width=0.235\linewidth]{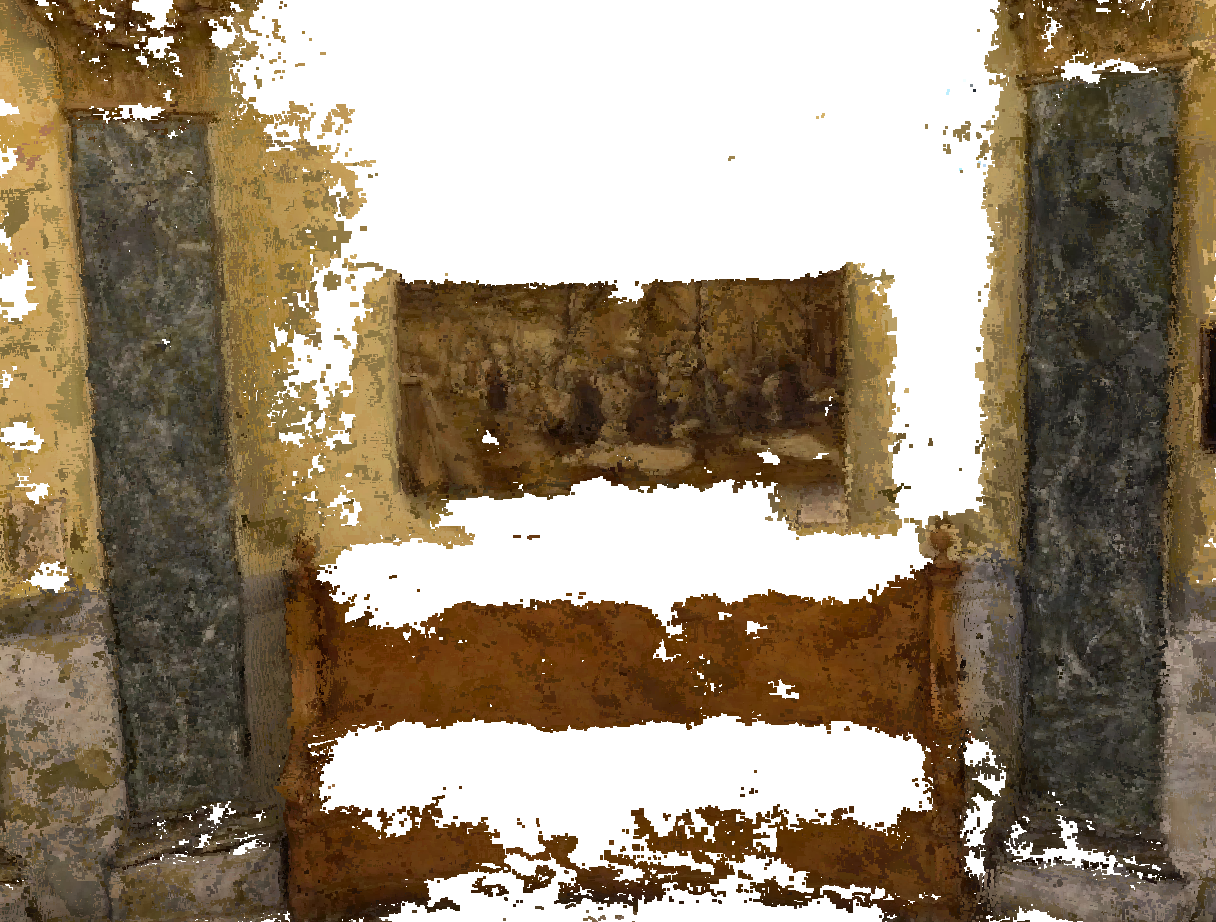} 
		\includegraphics[width=0.235\linewidth]{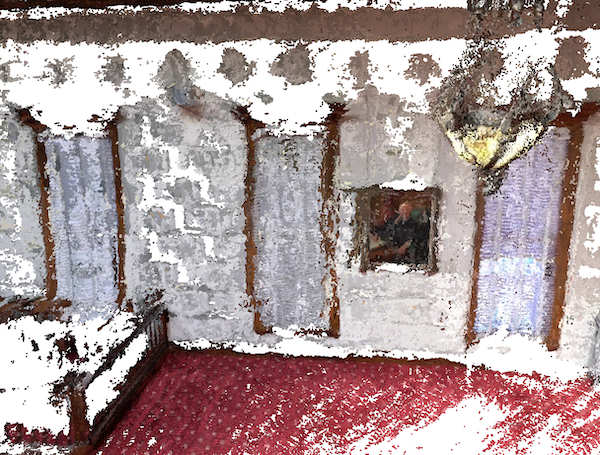} 
		\includegraphics[width=0.22\linewidth]{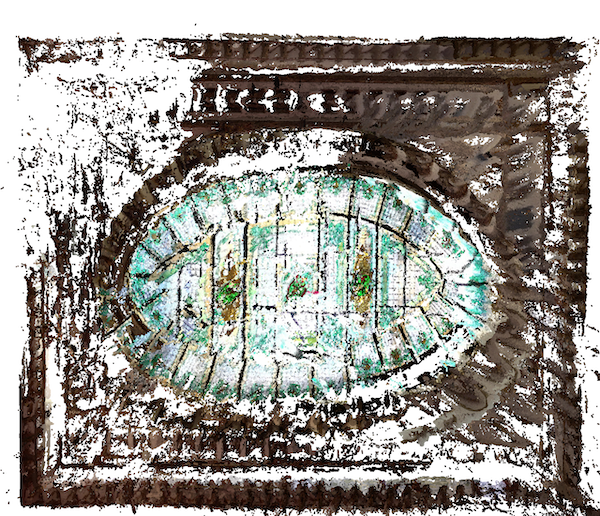}\\
		{PatchmatchNet (baseline)~\cite{wang2020patchmatchnet}}
		\label{subfig:PatchmatchNet}
	\end{tabular}
	\begin{tabular}{c@{}c@{}c@{}c@{}}
		\includegraphics[width=0.235\linewidth]{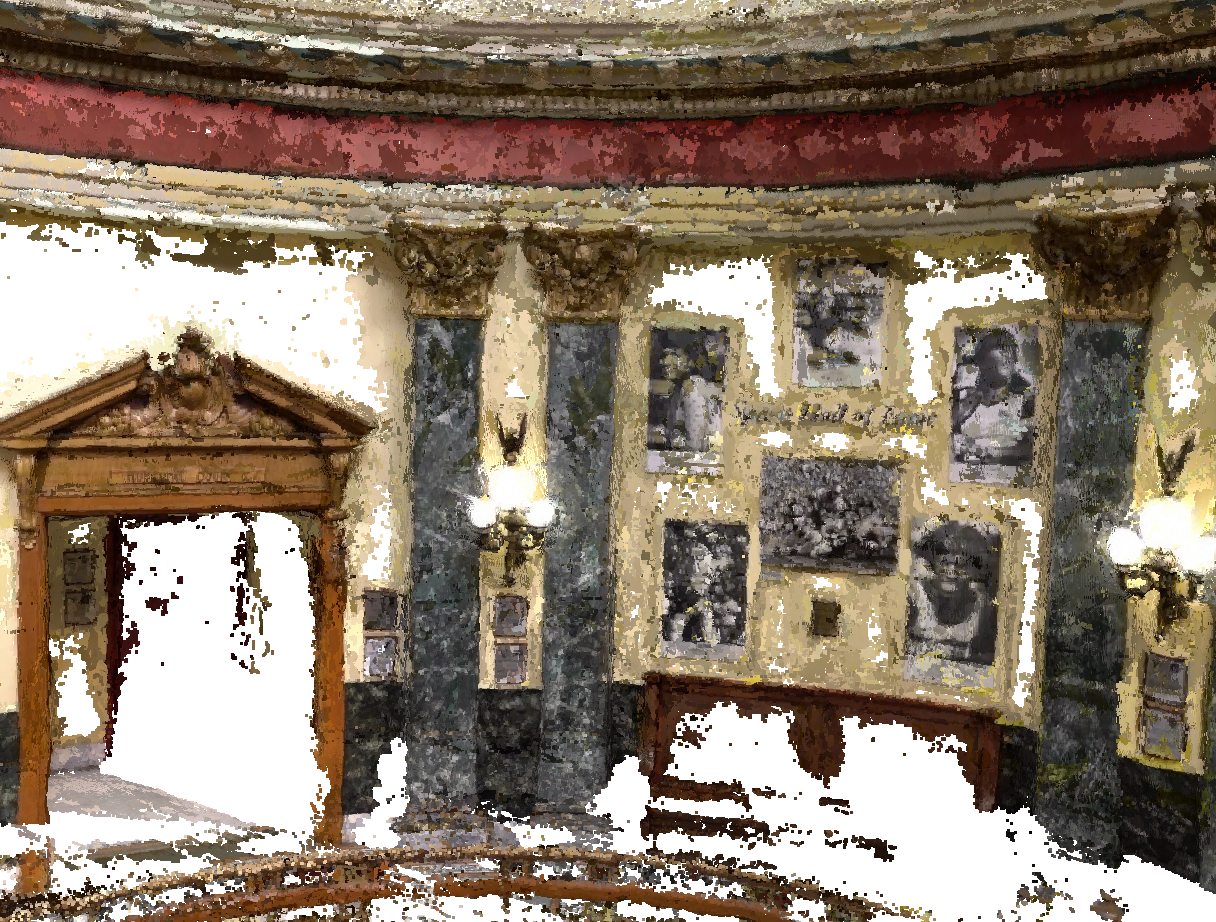} \hfill
		\includegraphics[width=0.235\linewidth]{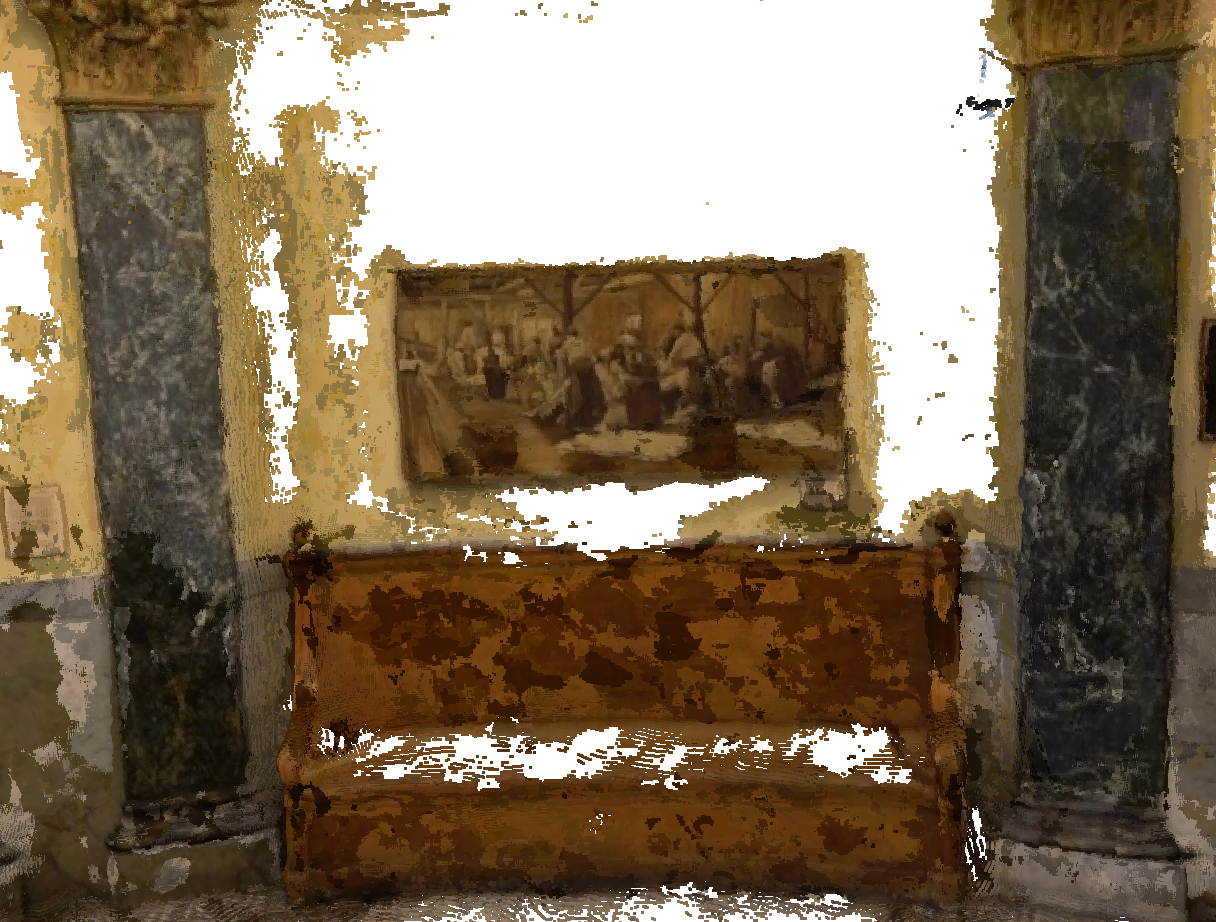} \hfill
		\includegraphics[width=0.235\linewidth]{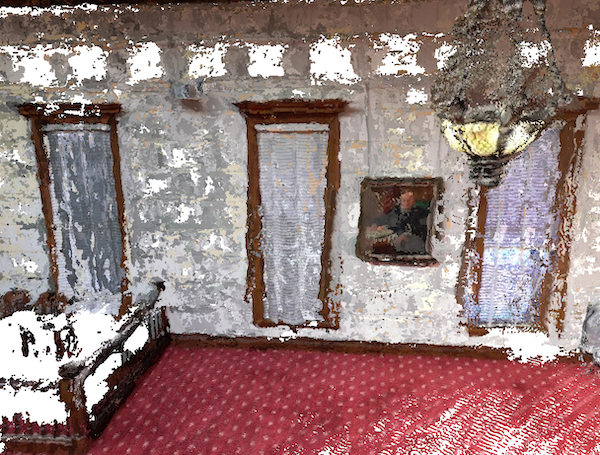}  \hfill
		\includegraphics[width=0.22\linewidth]{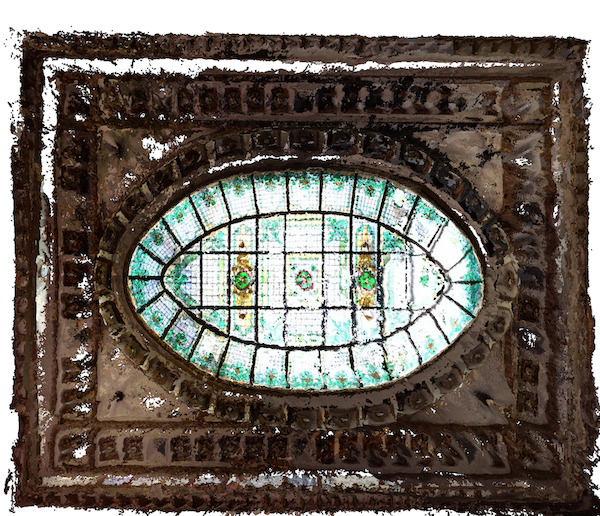}\\
		{Ours}
		\label{subfig:Ours}
	\end{tabular}
	\caption{Comparisons with the state-of-the-art traditional photogrammetry-based MVS method COLMAP~\cite{schoenberger2016mvs} and learning-based MVS method PatchmatchNet~\cite{wang2020patchmatchnet}.}
	\label{fig:compare}
\end{figure*}

Figure \ref{fig:compare} presents a comparison of our proposed learning-based MVS method with two  methods, namely COLMAP~\cite{schoenberger2016mvs} and PatchmatchNet~\cite{wang2020patchmatchnet} (our baseline). COLMAP is a state-of-the-art traditional photogrammetry-based method. We visualized the outcomes of the methods on four different scene parts from the ``Tanks and Temples''~\cite{Knapitsch2017} dataset, and the last column of the figure shows the exterior of the \textit{courtroom}'s top, with the lower part of the point cloud clipped to better reveal the ceiling's completeness and accuracy.

To ensure a fair comparison, we provided COLMAP with the ground-truth camera parameters. Our experiments demonstrate that our proposed method generates denser and more complete point clouds than the traditional photogrammetry-based method. However, the traditional method achieves better accuracy, partially due to its sparsity. Our method's results exhibit the highest completeness and are cleaner than the other methods. Additionally, our method outperforms PatchmatchNet~\cite{wang2020patchmatchnet} in terms of reconstruction accuracy. Please refer to the supplementary video for more visual comparisons.

\subsection{Generalization to aerial images}

To further evaluate the generalization capabilities of our proposed methods, we conducted experiments using aerial images. Aerial images are commonly used in remote sensing applications for tasks such as large-scale 3D reconstruction. For our experiments, we utilized the BlendedMVS dataset~\cite{yao2020blendedmvs}, which consists of aerial images with a low resolution of $768 \times 576$ images.

\begin{figure}[h]
	\centering
	\includegraphics[width=\textwidth]{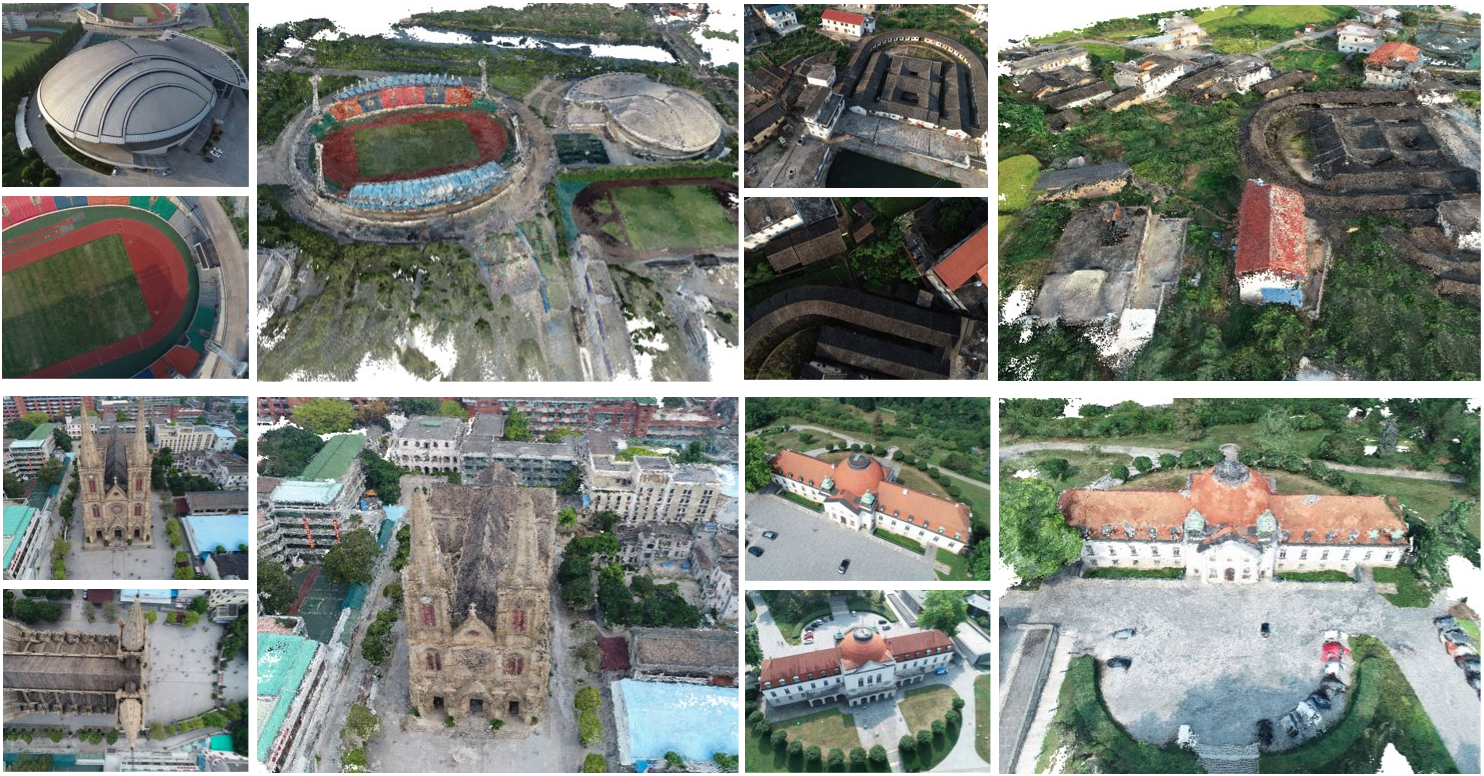}
	\caption{Experiments with BlendedMVS dataset. Qualitative results of our proposed methods for aerial image-based 3D reconstruction are visualized here.}
	\label{fig:aerial}
\end{figure}

Figure~\ref{fig:aerial} demonstrates some example images from the BlendedMVS dataset, illustrating the qualitative results obtained from our proposed methods. These results show that our method can generate  3D reconstructions from  aerial images, even with low-resolution input images. This indicates the potential of our approach for remote sensing applications that require large-scale 3D reconstructions from aerial imagery.

On a single RTX 2080, the time needed for depth inference per image is 90 $ms$ when using 5 neighboring views, and increases to 110 $ms$ when using 7 neighboring views. As an illustration of the running time, the bottom left building example in Fig. \ref{fig:aerial} is comprised of 77 images. It takes 79.993 seconds to generate a point cloud from the calibrated views with the default 5 neighboring views.

\subsection{Memory consumption and running times}
In Fig.~\ref{fig:edgemaps_memory}, we report our comparison of GPU memory demands with existing learning-based MVS networks on the DTU dataset~\cite{aanaes2016_dtu}, from which we can see that the memory demand of our network is much lower than most of the existing networks.  In the DTU dataset with the default parameters and the 5-view case, the average depth inference time for our model is 345$ms$. This is comparable to the performance of PatchmatchNet~\cite{wang2020patchmatchnet}, which took 300$ms$. We used a GPU of NVIDIA GeForce RTX 2080 for the experiments.

\subsection{Limitations}
Although our method has good completeness and a good overall score (see Tab.~\ref{tab:dtu_eval}), it has still not reached the accuracy level of traditional photogrammetry-based algorithms such as Gipuma~\cite{galliani_2015_gipuma}, which is a common weakness in recently developed learning-based MVS methods with high completeness score.
In this paper, our goal is to improve the accuracy of the reconstruction process while simultaneously maintaining a high level of completeness. Although the accuracy of our proposed network is not among the highest compared to some traditional state-of-the-art methods, we would like to emphasize that currently, learning-based approaches struggle to achieve a state-of-the-art accuracy result while maintaining a high completeness score. This is due to the trade-off between accuracy and completeness in the depth map fusion process, which is a key component of the reconstruction pipeline.
Such a trade-off implies that increasing completeness leads to an increasing potential source of noise. Although using bimodality helps to reduce the noise, we observe that our work, like other traditional and learning-based algorithms, contains noise, especially in sparsely viewed regions that may need further research. 
It is also worth noting that in this work we have used the same fusion pipeline as in other papers~\cite{wang2020patchmatchnet,yao_2018_mvsnet}.

%% file: source/conclusion.tex
\section{Conclusion}
\label{sec:conc}
We have presented a strategy for improving  the baseline MVS network by learning depth discontinuities.
The proposed depth discontinuity learning module has demonstrated superior performance compared to the baseline~\cite{wang2020patchmatchnet}. The results of our ablation study, as shown in Tab.~\ref{tab:ablation_dtutest_eval}, highlight the significant reduction in depth map error achieved by incorporating the proposed DDL module, reducing the error by more than 30\%. Experimental findings presented in Tab.~\ref{tab:boundary_region_dtu} demonstrate the enhanced quality of our approach in terms of depth map accuracy in smooth and boundary regions. Moreover,  our visual results shown in Fig.~\ref{fig:TNT_comparison} and Fig.~\ref{fig:DTU_boundary} revealed that the reconstructed point cloud obtained from our approach exhibits improved accuracy in capturing object and scene details  compared to the baseline model while maintaining completeness.

The results of Fig. \ref{fig:DTU_boundary} and Tab. \ref{tab:boundary_region_dtu} further reinforce the superiority of our method, with better qualitative and quantitative results in both smooth and boundary regions in the DTU~\cite{aanaes2016_dtu} dataset. These results indicate that our method has strong generalization capabilities and the ability to produce high-quality depth maps with improved accuracy and precision.  Furthermore, our experimental results demonstrate the potential of our method for remote sensing applications, such as large-scale point cloud reconstruction from aerial images.

Our experiments have demonstrated that learning depth maps as a mixture distribution and integrating depth discontinuities into the network as prior knowledge for piecewise smoothness regularization leads to improved reconstruction quality, with enhanced accuracy and overall quality of the final reconstruction.

%% file: main.bbl
\begin{thebibliography}{999}

\bibitem[Lemaire(2008)]{lemaire2008aspects}
Lemaire, C.
\newblock Aspects of the DSM production with high resolution images.
\newblock {\em ISPRS} {\bf 2008}, {\em 37},~1143--1146.

\bibitem[Peppa \em{et~al.}(2019)Peppa, Mills, Moore, Miller, and
  Chambers]{peppa2019automated}
Peppa, M.V.; Mills, J.P.; Moore, P.; Miller, P.E.; Chambers, J.E.
\newblock Automated co-registration and calibration in SfM photogrammetry for
  landslide change detection.
\newblock {\em Earth Surf. Process. Landf.} {\bf 2019}, {\em 44},~287--303.

\bibitem[Nguatem and Mayer(2017)]{Nguatem_2017_ICCV}
Nguatem, W.; Mayer, H.
\newblock Modeling urban scenes from pointclouds.
\newblock  ICCV. IEEE,  2017, pp. 3857--3866.

\bibitem[Furukawa and Ponce(2010)]{furukawa2010}
Furukawa, Y.; Ponce, J.
\newblock Accurate, dense, and robust multi-view stereopsis.
\newblock {\em IEEE TPAMI} {\bf 2010}, {\em 32},~1362--1376.

\bibitem[Galliani \em{et~al.}(2015)Galliani, Lasinger, and
  Schindler]{galliani_2015_gipuma}
Galliani, S.; Lasinger, K.; Schindler, K.
\newblock Massively parallel multiview stereopsis by surface normal diffusion.
\newblock  ICCV. IEEE,  2015, pp. 873--881.

\bibitem[Tola \em{et~al.}(2012)Tola, Strecha, and Fua]{tola2012efficient}
Tola, E.; Strecha, C.; Fua, P.
\newblock Efficient large-scale multi-view stereo for ultra high-resolution
  image sets.
\newblock {\em Machine vision and applications} {\bf 2012}, {\em 23},~903--920.

\bibitem[Yao \em{et~al.}(2018)Yao, Luo, Li, Fang, and Quan]{yao_2018_mvsnet}
Yao, Y.; Luo, Z.; Li, S.; Fang, T.; Quan, L.
\newblock MVSNet: Depth inference for unstructured multi-view stereo.
\newblock  ECCV. Springer,  2018, pp. 767--783.

\bibitem[Yao \em{et~al.}(2019)Yao, Luo, Li, Shen, Fang, and
  Quan]{yao_2019_rmvsnet}
Yao, Y.; Luo, Z.; Li, S.; Shen, T.; Fang, T.; Quan, L.
\newblock Recurrent MVSNet for high-resolution multi-view stereo depth
  inference.
\newblock  CVPR. IEEE,  2019, pp. 5525--5534.

\bibitem[Ji \em{et~al.}(2017)Ji, Gall, Zheng, Liu, and
  Fang]{ji_2017_surfacenet}
Ji, M.; Gall, J.; Zheng, H.; Liu, Y.; Fang, L.
\newblock {SurfaceNet:} An end-to-end 3D neural network for multiview
  stereopsis.
\newblock  ICCV. IEEE,  2017, pp. 2307--2315.

\bibitem[Chen \em{et~al.}(2019)Chen, Han, Xu, and Su]{chen_2019_pointmvsnet}
Chen, R.; Han, S.; Xu, J.; Su, H.
\newblock Point-based multi-view stereo network.
\newblock  ICCV,  2019, pp. 1538--1547.

\bibitem[Yu and Gao(2020)]{Yu_2020_fastmvsnet}
Yu, Z.; Gao, S.
\newblock Fast-MVSNet: Sparse-to-dense multi-view stereo with learned
  propagation and Gauss-Newton refinement.
\newblock  CVPR. IEEE,  2020, pp. 1949--1958.

\bibitem[Cheng \em{et~al.}(2020)Cheng, Xu, Zhu, Li, Li, Ramamoorthi, and
  Su]{cheng_2020_ucsnet}
Cheng, S.; Xu, Z.; Zhu, S.; Li, Z.; Li, L.E.; Ramamoorthi, R.; Su, H.
\newblock Deep stereo using adaptive thin volume representation with
  uncertainty awareness.
\newblock  CVPR. IEEE,  2020, pp. 2524--2534.

\bibitem[Gu \em{et~al.}(2020)Gu, Fan, Zhu, Dai, Tan, and
  Tan]{gu_2020_cascademvsnet}
Gu, X.; Fan, Z.; Zhu, S.; Dai, Z.; Tan, F.; Tan, P.
\newblock Cascade cost volume for high-resolution multi-view stereo and stereo
  matching.
\newblock  CVPR. IEEE,  2020, pp. 2495--2504.

\bibitem[Luo \em{et~al.}(2019)Luo, Guan, Ju, Huang, and Luo]{luo_2019_p_mvsnet}
Luo, K.; Guan, T.; Ju, L.; Huang, H.; Luo, Y.
\newblock P-MVSNet: Learning patch-wise matching confidence aggregation for
  multi-view stereo.
\newblock  ICCV. IEEE,  2019, pp. 10451--10460.

\bibitem[Xu and Tao(2020)]{xu2020learning_inverse}
Xu, Q.; Tao, W.
\newblock Learning inverse depth regression for multi-view stereo with
  correlation cost volume.
\newblock  AAAI,  2020, Vol.~34, pp. 12508--12515.

\bibitem[Yang \em{et~al.}(2020)Yang, Mao, Alvarez, and Liu]{yang_2020_cvpr}
Yang, J.; Mao, W.; Alvarez, J.M.; Liu, M.
\newblock Cost volume pyramid based depth inference for multi-view stereo.
\newblock  CVPR. IEEE,  2020, pp. 4877--4886.

\bibitem[Wang \em{et~al.}(2021)Wang, Galliani, Vogel, Speciale, and
  Pollefeys]{wang2020patchmatchnet}
Wang, F.; Galliani, S.; Vogel, C.; Speciale, P.; Pollefeys, M.
\newblock Patchmatchnet: Learned multi-view patchmatch stereo.
\newblock  CVPR. IEEE,  2021, pp. 14194--14203.

\bibitem[Duggal \em{et~al.}(2019)Duggal, Wang, Ma, Hu, and
  Urtasun]{Duggal2019ICCV}
Duggal, S.; Wang, S.; Ma, W.C.; Hu, R.; Urtasun, R.
\newblock DeepPruner: Learning efficient stereo matching via differentiable
  patchmatch.
\newblock  ICCV. IEEE,  2019, pp. 4384--4393.

\bibitem[Zhu \em{et~al.}(2020)Zhu, Brazil, and Liu]{zhu2020edge}
Zhu, S.; Brazil, G.; Liu, X.
\newblock The edge of depth: Explicit constraints between segmentation and
  depth.
\newblock  CVPR. IEEE,  2020, pp. 13116--13125.

\bibitem[Tosi \em{et~al.}(2021)Tosi, Liao, Schmitt, and Geiger]{Tosi2021CVPR}
Tosi, F.; Liao, Y.; Schmitt, C.; Geiger, A.
\newblock SMD-Nets: Stereo mixture density networks.
\newblock  CVPR. IEEE,  2021, pp. 8942--8952.

\bibitem[Boykov \em{et~al.}(2001)Boykov, Veksler, and Zabih]{boykov2001fast}
Boykov, Y.; Veksler, O.; Zabih, R.
\newblock Fast approximate energy minimization via graph cuts.
\newblock {\em IEEE TPAMI} {\bf 2001}, {\em 23},~1222--1239.

\bibitem[Boykov \em{et~al.}(1998)Boykov, Veksler, and Zabih]{boykov_cvpr_1998}
Boykov, Y.; Veksler, O.; Zabih, R.
\newblock Markov random fields with efficient approximations.
\newblock  CVPR. IEEE,  1998, pp. 648--655.

\bibitem[Garg \em{et~al.}(2020)Garg, Wang, Hariharan, Campbell, Weinberger, and
  Chao]{garg2020wasserstein}
Garg, D.; Wang, Y.; Hariharan, B.; Campbell, M.; Weinberger, K.Q.; Chao, W.L.
\newblock Wasserstein distances for stereo disparity estimation.
\newblock  NeurIPS,  2020, Vol.~33, pp. 22517--22529.

\bibitem[Janai \em{et~al.}(2020)Janai, G{\"u}ney, Behl, Geiger,
  et~al.]{janai2020computer}
Janai, J.; G{\"u}ney, F.; Behl, A.; Geiger, A.;  et~al.
\newblock Computer vision for autonomous vehicles: Problems, datasets and state
  of the art.
\newblock {\em Foundations and Trends{\textregistered} in Computer Graphics and
  Vision} {\bf 2020}, {\em 12},~1--308.

\bibitem[Kutulakos and Seitz(2000)]{kutulakos2000theory}
Kutulakos, K.N.; Seitz, S.M.
\newblock A theory of shape by space carving.
\newblock {\em IJCV} {\bf 2000}, {\em 38},~199--218.

\bibitem[Faugeras and Keriven(2002)]{faugeras2002variational}
Faugeras, O.; Keriven, R.
\newblock {\em Variational principles, surface evolution, PDE's, level set
  methods and the stereo problem}; IEEE,  2002.

\bibitem[Lorensen and Cline(1987)]{lorensen1987marching}
Lorensen, W.E.; Cline, H.E.
\newblock Marching cubes: A high resolution 3D surface construction algorithm.
\newblock {\em ACM siggraph computer graphics} {\bf 1987}, {\em 21},~163--169.

\bibitem[Curless and Levoy(1996)]{curless1996volumetric}
Curless, B.; Levoy, M.
\newblock A volumetric method for building complex models from range images.
\newblock  Computer graphics and interactive techniques,  1996, pp. 303--312.

\bibitem[Zach \em{et~al.}(2007)Zach, Pock, and Bischof]{zach2007globally}
Zach, C.; Pock, T.; Bischof, H.
\newblock A globally optimal algorithm for robust tv-l 1 range image
  integration.
\newblock  ICCV. IEEE,  2007, pp. 1--8.

\bibitem[Collins(1996)]{collins1996space}
Collins, R.T.
\newblock A space-sweep approach to true multi-image matching.
\newblock  CVPR. IEEE,  1996, pp. 358--363.

\bibitem[Pollefeys \em{et~al.}(2008)Pollefeys, Nist{\'e}r, Frahm, Akbarzadeh,
  Mordohai, Clipp, Engels, Gallup, Kim, Merrell, et~al.]{pollefeys2008detailed}
Pollefeys, M.; Nist{\'e}r, D.; Frahm, J.M.; Akbarzadeh, A.; Mordohai, P.;
  Clipp, B.; Engels, C.; Gallup, D.; Kim, S.J.; Merrell, P.;  et~al.
\newblock Detailed real-time urban 3d reconstruction from video.
\newblock {\em IJCV} {\bf 2008}, {\em 78},~143--167.

\bibitem[Sch\"{o}nberger \em{et~al.}(2016)Sch\"{o}nberger, Zheng, Pollefeys,
  and Frahm]{schoenberger2016mvs}
Sch\"{o}nberger, J.L.; Zheng, E.; Pollefeys, M.; Frahm, J.M.
\newblock Pixelwise view selection for unstructured multi-view stereo.
\newblock  ECCV. Springer,  2016.

\bibitem[Zbontar \em{et~al.}(2016)Zbontar, LeCun, et~al.]{zbontar2016stereo}
Zbontar, J.; LeCun, Y.;  et~al.
\newblock Stereo matching by training a convolutional neural network to compare
  image patches.
\newblock {\em J. Mach. Learn. Res.} {\bf 2016}, {\em 17},~2287--2318.

\bibitem[Kendall \em{et~al.}(2017)Kendall, Martirosyan, Dasgupta, Henry,
  Kennedy, Bachrach, and Bry]{kendall2017end}
Kendall, A.; Martirosyan, H.; Dasgupta, S.; Henry, P.; Kennedy, R.; Bachrach,
  A.; Bry, A.
\newblock End-to-end learning of geometry and context for deep stereo
  regression.
\newblock  ICCV. IEEE,  2017, pp. 66--75.

\bibitem[Chang and Chen(2018)]{chang2018pyramid}
Chang, J.R.; Chen, Y.S.
\newblock Pyramid stereo matching network.
\newblock  CVPR. IEEE,  2018, pp. 5410--5418.

\bibitem[Yang \em{et~al.}(2019)Yang, Manela, Happold, and
  Ramanan]{yang2019hierarchical}
Yang, G.; Manela, J.; Happold, M.; Ramanan, D.
\newblock Hierarchical deep stereo matching on high-resolution images.
\newblock  CVPR. IEEE,  2019, pp. 5515--5524.

\bibitem[Song \em{et~al.}(2018)Song, Zhao, Hu, and Fang]{song2018edgestereo}
Song, X.; Zhao, X.; Hu, H.; Fang, L.
\newblock Edgestereo: A context integrated residual pyramid network for stereo
  matching.
\newblock  ACCV. Springer,  2018.

\bibitem[Lin \em{et~al.}(2021)Lin, Li, Zhang, Zheng, and Wu]{lin2021high}
Lin, K.; Li, L.; Zhang, J.; Zheng, X.; Wu, S.
\newblock High-Resolution Multi-View Stereo with Dynamic Depth Edge Flow.
\newblock  ICME. IEEE,  2021, pp. 1--6.

\bibitem[Ding \em{et~al.}(2022)Ding, Li, Huang, Zhang, Li, and
  Feng]{ding2022adaptive}
Ding, Y.; Li, Z.; Huang, D.; Zhang, K.; Li, Z.; Feng, W.
\newblock Adaptive Range guided Multi-view Depth Estimation with Normal Ranking
  Loss.
\newblock  ACCV,  2022, pp. 1892--1908.

\bibitem[Zhang \em{et~al.}(2022)Zhang, Tang, Cao, Xiao, Huang, and
  Fang]{zhang2022elasticmvs}
Zhang, J.; Tang, R.; Cao, Z.; Xiao, J.; Huang, R.; Fang, L.
\newblock ElasticMVS: Learning elastic part representation for self-supervised
  multi-view stereopsis.
\newblock {\em NeurIPS} {\bf 2022}, {\em 35},~23510--23523.

\bibitem[Zhang \em{et~al.}(2023)Zhang, Yang, Chang, and Qin]{zhang2023mg}
Zhang, X.; Yang, F.; Chang, M.; Qin, X.
\newblock MG-MVSNet: Multiple Granularities Feature Fusion Network for
  Multi-View Stereo.
\newblock {\em Neurocomputing} {\bf 2023}.

\bibitem[Ronneberger \em{et~al.}(2015)Ronneberger, Fischer, and
  Brox]{ronneberger2015u}
Ronneberger, O.; Fischer, P.; Brox, T.
\newblock U-net: Convolutional networks for biomedical image segmentation.
\newblock  Medical Image Computing and Computer-Assisted Intervention.
  Springer,  2015, pp. 234--241.

\bibitem[Lin \em{et~al.}(2017)Lin, Doll{\'a}r, Girshick, He, Hariharan, and
  Belongie]{lin2017feature}
Lin, T.Y.; Doll{\'a}r, P.; Girshick, R.; He, K.; Hariharan, B.; Belongie, S.
\newblock Feature pyramid networks for object detection.
\newblock  CVPR. IEEE,  2017, pp. 2117--2125.

\bibitem[Hui \em{et~al.}(2016)Hui, Loy, and Tang]{hui16msgnet}
Hui, T.W.; Loy, C.C.; Tang, X.
\newblock Depth map super-resolution by deep multi-scale guidance.
\newblock  ECCV. Springer,  2016.

\bibitem[Yu \em{et~al.}(2021)Yu, Guo, Liu, Chen, Wang, Cao, and
  Jiang]{yu2021attention}
Yu, A.; Guo, W.; Liu, B.; Chen, X.; Wang, X.; Cao, X.; Jiang, B.
\newblock Attention aware cost volume pyramid based multi-view stereo network
  for 3D reconstruction.
\newblock {\em ISPRS J. of Photogrammetry and Remote Sensing} {\bf 2021}, {\em
  175},~448--460.

\bibitem[Huang \em{et~al.}(2000)Huang, Lee, and Mumford]{huang00}
Huang, J.; Lee, A.; Mumford, D.
\newblock Statistics of range images.
\newblock  CVPR. IEEE,  2000, Vol.~1, pp. 324--331 vol.1.

\bibitem[Laplace(1801)]{laplace_distribution}
Laplace, P.S.
\newblock Laplace distribution.
\newblock {\em Encyclopedia of Mathematics} {\bf 1801}.
\newblock Original publication in 1801, available in English translation.

\bibitem[Aan{\ae}s \em{et~al.}(2016)Aan{\ae}s, Jensen, Vogiatzis, Tola, and
  Dahl]{aanaes2016_dtu}
Aan{\ae}s, H.; Jensen, R.R.; Vogiatzis, G.; Tola, E.; Dahl, A.B.
\newblock Large-scale data for multiple-view stereopsis.
\newblock {\em IJCV} {\bf 2016}, pp. 1--16.

\bibitem[Laplace(1820)]{laplace_operator}
Laplace, P.S.
\newblock Laplace operator.
\newblock {\em Encyclopedia of Mathematics} {\bf 1820}.
\newblock Original publication in 1820, available in English translation.

\bibitem[Knapitsch \em{et~al.}(2017)Knapitsch, Park, Zhou, and
  Koltun]{Knapitsch2017}
Knapitsch, A.; Park, J.; Zhou, Q.Y.; Koltun, V.
\newblock Tanks and Temples: Benchmarking large-scale scene reconstruction.
\newblock {\em ACM TOG} {\bf 2017}, {\em 36}.

\bibitem[Sch\"ops \em{et~al.}(2017)Sch\"ops, Sch\"onberger, Galliani, Sattler,
  Schindler, Pollefeys, and Geiger]{schoeps2017cvpr}
Sch\"ops, T.; Sch\"onberger, J.L.; Galliani, S.; Sattler, T.; Schindler, K.;
  Pollefeys, M.; Geiger, A.
\newblock A Multi-View Stereo Benchmark with high-resolution images and
  multi-camera videos.
\newblock  CVPR. IEEE,  2017.

\bibitem[Yao \em{et~al.}(2020)Yao, Luo, Li, Zhang, Ren, Zhou, Fang, and
  Quan]{yao2020blendedmvs}
Yao, Y.; Luo, Z.; Li, S.; Zhang, J.; Ren, Y.; Zhou, L.; Fang, T.; Quan, L.
\newblock Blendedmvs: A large-scale dataset for generalized multi-view stereo
  networks.
\newblock  CVPR. IEEE,  2020, pp. 1790--1799.

\bibitem[Peng \em{et~al.}(2022)Peng, Wang, Wang, Lai, and
  Wang]{peng2022rethinking}
Peng, R.; Wang, R.; Wang, Z.; Lai, Y.; Wang, R.
\newblock Rethinking depth estimation for multi-view stereo: A unified
  representation.
\newblock  CVPR. IEEE,  2022, pp. 8645--8654.

\bibitem[Wang \em{et~al.}(2022)Wang, Zhu, Huang, Qin, Ye, He, Chi, and
  Wang]{wang2022mvster}
Wang, X.; Zhu, Z.; Huang, G.; Qin, F.; Ye, Y.; He, Y.; Chi, X.; Wang, X.
\newblock MVSTER: epipolar transformer for efficient multi-view stereo.
\newblock  ECCV. Springer,  2022, pp. 573--591.

\bibitem[Campbell \em{et~al.}(2008)Campbell, Vogiatzis, Hern{\'a}ndez, and
  Cipolla]{camp}
Campbell, N.D.F.; Vogiatzis, G.; Hern{\'a}ndez, C.; Cipolla, R.
\newblock Using Multiple Hypotheses to Improve Depth-Maps for Multi-View
  Stereo.
\newblock  ECCV; Forsyth, D.; Torr, P.; Zisserman, A., Eds.; Springer Berlin
  Heidelberg: Berlin, Heidelberg,  2008; pp. 766--779.

\bibitem[Luo \em{et~al.}(2020)Luo, Guan, Ju, Wang, Chen, and
  Luo]{Luo_2020_CVPR}
Luo, K.; Guan, T.; Ju, L.; Wang, Y.; Chen, Z.; Luo, Y.
\newblock Attention-aware multi-view stereo.
\newblock  CVPR. IEEE,  2020.

\bibitem[Zhang \em{et~al.}(2020)Zhang, Yao, Li, Luo, and
  Fang]{zhang2020visibility}
Zhang, J.; Yao, Y.; Li, S.; Luo, Z.; Fang, T.
\newblock Visibility-aware multi-view stereo network.
\newblock  BMVC,  2020.

\bibitem[Ma \em{et~al.}(2021)Ma, Gong, Wang, Huang, Chen, and Yu]{ma2021epp}
Ma, X.; Gong, Y.; Wang, Q.; Huang, J.; Chen, L.; Yu, F.
\newblock EPP-MVSNet: Epipolar-assembling based depth prediction for multi-view
  stereo.
\newblock  ICCV. IEEE,  2021, pp. 5732--5740.

\bibitem[Wei \em{et~al.}(2021)Wei, Zhu, Min, Chen, and Wang]{wei2021aa}
Wei, Z.; Zhu, Q.; Min, C.; Chen, Y.; Wang, G.
\newblock AA-RMVSNet: Adaptive aggregation recurrent multi-view stereo network.
\newblock  ICCV. IEEE,  2021, pp. 6187--6196.

\bibitem[Xie and Tu(2015)]{xie2015holistically}
Xie, S.; Tu, Z.
\newblock Holistically-nested edge detection.
\newblock  ICCV. IEEE,  2015, pp. 1395--1403.

\end{thebibliography}
